\documentclass[letterpaper, 10 pt, journal, twoside]{ieeetran}
\UseRawInputEncoding
\usepackage[ruled,vlined,linesnumbered]{algorithm2e}
\usepackage{amsmath,amsfonts}
\usepackage{caption}
\usepackage{booktabs}
\usepackage{multirow}
\usepackage{makecell}
\usepackage{diagbox}
\usepackage{array}
\usepackage{booktabs}
\usepackage{bigstrut}
\usepackage{color}
\usepackage{xcolor}
\usepackage{colortbl}
\usepackage[colorlinks,
            linkcolor=red,
            anchorcolor=red,
            citecolor=red]{hyperref}
            
\usepackage{authblk}
\usepackage{array}
\usepackage[caption=false,font=normalsize,labelfont=sf,textfont=sf]{subfig}
\usepackage{textcomp}
\usepackage{stfloats}
\usepackage{url}
\usepackage{verbatim}
\usepackage{graphicx}
\usepackage{cite}
\usepackage{bbm}
\usepackage{amsmath}

\definecolor{mygreen}{rgb}{.75,1,.75}
\usepackage{etoolbox}
\makeatletter
\patchcmd{\@makecaption}
{\scshape}
{}
{}
{}
\newcommand{\customref}[2]{\hyperref[#2]{#1}}
\usepackage{pifont}

\usepackage[capitalize]{cleveref}
\makeatother

\begin{document}

\title{Offline-Poly: A Polyhedral Framework For Offline 3D Multi-Object Tracking}

\author{Xiaoyu Li, Yitao Wu, Xian Wu, Haolin Zhuo, Lijun Zhao,~\IEEEmembership{Member,~IEEE,} Lining Sun
\thanks{This work was supported by the Self-Planned Task of State Key Laboratory of Robotics (SKLRS202501E). (Yitao Wu and Xian Wu contributed equally. 
Corresponding author: Lijun Zhao.)}
\thanks{All authors are with the State Key Laboratory of Robotics and System, Harbin Institute of Technology, Harbin 150006, China.}
}

\markboth{This work has been submitted to the IEEE for possible publication.}%
{Li \MakeLowercase{\textit{et al.}}: Offline-Poly: A Polyhedral Framework For Offline 3D Multi-Object Tracking}


\maketitle

\begin{abstract}

Offline 3D multi-object tracking (MOT) is a critical component of the 4D auto-labeling (4DAL) process for dynamic obstacles. 
It enhances pseudo-labels generated by high-performance detectors through the incorporation of temporal context.
However, most existing offline 3D MOT approaches are direct extensions of online Tracking-By-Detection (TBD) frameworks and fail to fully exploit the advantages of the offline setting.
Moreover, dedicated offline trackers often depend on fixed upstream inputs and customized architectures, limiting their adaptability across diverse scenarios.
To address these limitations, we propose Offline-Poly, a general and flexible offline 3D MOT method based on a tracking-centric design.
We introduce a standardized paradigm termed Tracking-by-Tracking (TBT), which operates exclusively on arbitrary off-the-shelf tracking outputs and produces offline-refined trajectories.
This formulation decouples the offline tracker from specific upstream detectors or trackers, ensuring strong generalization capability.
Under the TBT paradigm, Offline-Poly accepts one or multiple coarse tracking results from arbitrary upstream trackers and processes them through a structured pipeline comprising pre-processing, hierarchical matching and fusion, and trajectory refinement.
Each module is deliberately designed to capitalize on the two fundamental properties of offline tracking: resource unconstrainedness, which permits global optimization beyond real-time limits, and future observability, which enables trajectory reasoning over the full temporal horizon.
Specifically, Offline-Poly first eliminates short-term ghost tracklets and re-identifies fragmented segments using global scene context.
It then constructs scene-level similarity to associate tracklets across multiple input sources, followed by a heuristic fusion strategy. 
Finally, Offline-Poly refines trajectories by jointly leveraging local and global motion patterns.
We evaluate Offline-Poly extensively on two autonomous driving datasets.
On the nuScenes dataset, Offline-Poly achieves state-of-the-art performance with 77.6\% AMOTA.
On the KITTI dataset, it achieves leading results with 83.00\% HOTA.
Comprehensive ablation studies further validate the flexibility, generalizability, and modular effectiveness of Offline-Poly.
Code will be released.

\end{abstract}


\section{Introduction}

4D auto-labeling (4DAL) system efficiently and economically generates high-quality pseudo ground-truth, forming a cornerstone of data-driven end-to-end autonomous driving (E2E AD) systems~\cite{e2esurvey}.
Within this pipeline, offline 3D multi-object tracking (MOT) plays a critical role in annotating dynamic objects~\cite{ma2023detzero, CTRL, 3DAL, Auto4D}.
It aggregates frame-level detections into instance-level tracklets, bridging the gap between per-frame detection outputs and temporally consistent 4D annotations.

Despite its importance, most offline 3D MOT implementations remain adaptations of online \textit{Tracking-By-Detection (TBD)} trackers~\cite{weng20203d, wang2021immortal, pang2022simpletrack}, with only minor modifications for offline usage.
Common strategies (weighted box fusion~\cite{ma2023detzero, WBF}, extending tracklet age~\cite{wang2021immortal, CTRL}, bidirectional tracking~\cite{ma2023detzero, CTRL}, etc.) are heuristically incorporated into online pipelines to approximate offline behavior.
Recently, offline 3D MOT has emerged as an independent research problem, leading to dedicated approaches~\cite{OTOP, RoboMOT, bitrack} that more explicitly exploit full-sequence information.
These approaches enhance trajectory quality through segment re-identification~\cite{OTOP, RoboMOT}, outlier filtering~\cite{RoboMOT}, and bidirectional integration~\cite{bitrack}.
However, most existing offline trackers encode offline reasoning within specialized architectures tightly coupled to specific inputs, which limits their generalizability and flexibility.
More importantly, they often exploit only isolated aspects of the offline paradigm.
As shown in Fig.~\ref{fig:tease}, we argue that offline tracking is fundamentally characterized by two intrinsic properties that distinguish it from online tracking:


\begin{figure}
    \centering
    \includegraphics[width=1.0\linewidth]{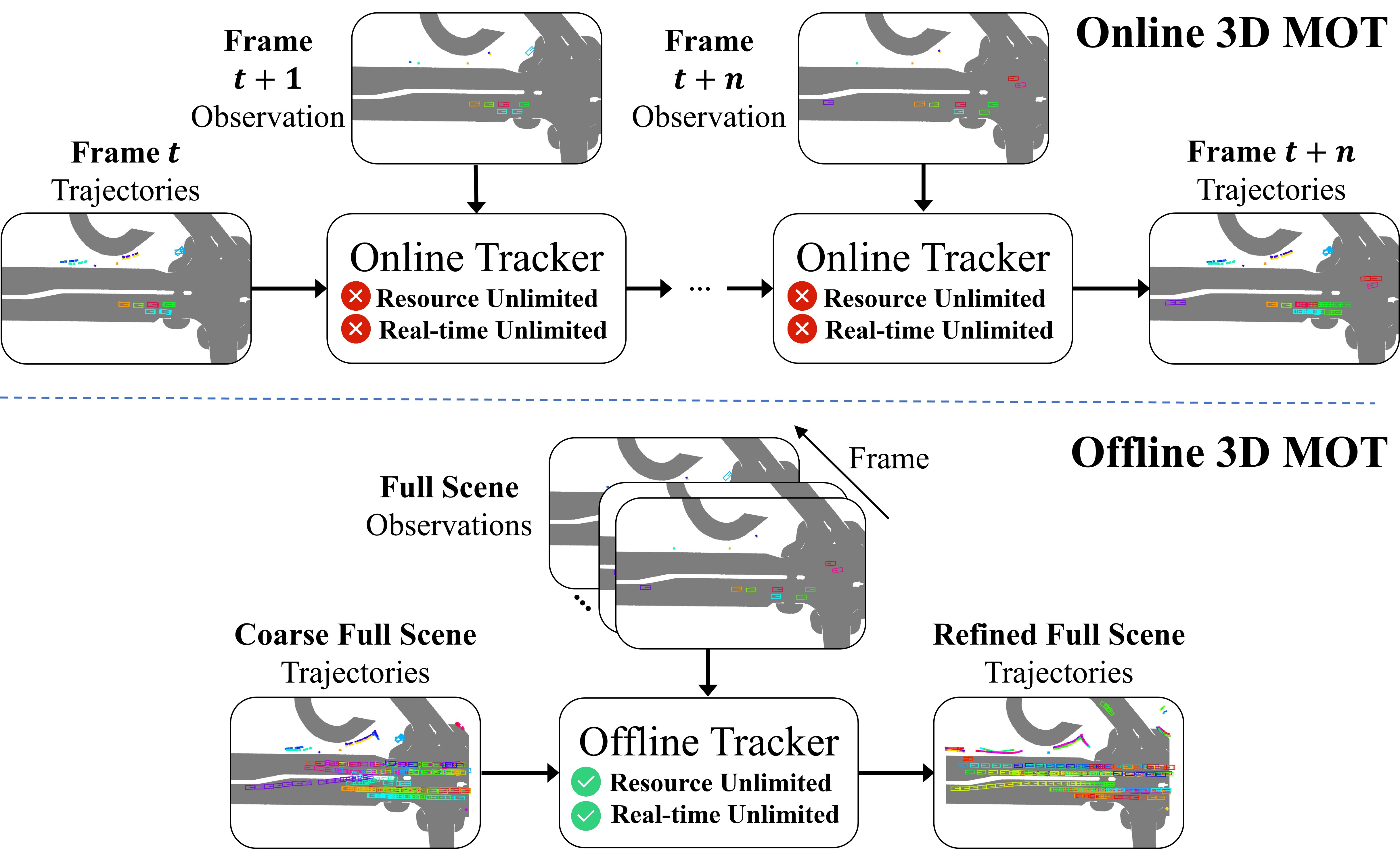}
    \caption{The comparison between online and offline 3D MOT. Both methods produce object trajectories, but offline 3D MOT can exploit the complete scene information and is not constrained by real-time requirements.}
    \label{fig:tease}
    \vspace{-1.5em}
\end{figure}


\textbf{Resource Unconstrainedness.}
Offline tracking is not bound by real-time constraints, enabling computationally intensive global reasoning and structural reorganization.
Nevertheless, most existing methods operate under a single-in-single-out (SISO) paradigm, consuming either a single detection stream~\cite{CTRL, 3DAL, Auto4D} or a single tracking result~\cite{RoboMOT, PC3T, wang2025mctrack}.
Even methods that merge forward and backward passes~\cite{ma2023detzero, bitrack} remain tied to a single detection source.
We observe that the absence of resource constraints naturally permits a multiple-input-single-output (MISO) formulation, where various independent tracking hypotheses can be jointly consolidated. 
Under this paradigm, complementary observations from distinct trackers can be fused to produce more complete object evidence, enhancing robustness and annotation quality.


\textbf{Future Observability.}
Offline tracking has access to the entire temporal sequence, enabling future-guided reasoning.
Some approaches leverage this property in specific forms, such as detecting outliers based on overall trajectory trends~\cite{RoboMOT}, or assessing trajectory similarity via map and geometry priors~\cite{OTOP}. 
However, these mechanisms are typically embedded within tailored model designs, restricting their applicability.
We argue that future observability should be incorporated as a general design principle within an input-agnostic framework.
Full-sequence visibility enables reliable tracklet filtering during pre-processing, robust affinity estimation during matching, consistent multi-source alignment during fusion, and motion smoothing through future-aware refinement.

Building on these insights, we propose Offline-Poly, a polyhedral framework for offline 3D MOT grounded in a novel \textit{Tracking-By-Tracking (TBT)} paradigm.
TBT treats upstream tracking results as atomic inputs and produces globally refined trajectories optimized for the offline setting.
In contrast to offline trackers that rely on frame-wise detections or intermediate online association metadata, TBT requires only final tracklets from arbitrary upstream trackers, decoupling the offline module from specific detector or tracker architectures.
Under the TBT paradigm, Offline-Poly unifies intra- and inter-tracker consolidation within a tracking-centric pipeline consisting of four modular components.
Each module is explicitly designed to leverage resource unconstrainedness and future observability.
(\uppercase\expandafter{\romannumeral1}) \textbf{Pre-processing:}
Low-quality trajectories are filtered using lifecycle and confidence statistics, effectively eliminating short-lived ghost tracklets under global sequence awareness.
(\uppercase\expandafter{\romannumeral2}) \textbf{Matching} and (\uppercase\expandafter{\romannumeral3}) \textbf{Fusion:}
We then introduce a novel hierarchical matching and fusion module to consolidate tracklets both within individual trackers and across multiple trackers.
Full-sequence visibility enables reliable association of fragmented segments and resolution of identity ambiguities.
Within each tracker, tracklets are linked and reorganized via geometry-based affinity construction and solving.
Cross-tracker association clusters spatially similar tracklets based on scene-level similarity, followed by fusion to exploit complementary evidence.
(\uppercase\expandafter{\romannumeral4}) \textbf{Refinement:} 
A multi-perspective refinement module performs geometric and kinematic optimization over complete trajectories. 
We apply a corner-alignment strategy to globally smooth trajectories under object rigidity constraints, followed by a local sliding-window optimization that refines motion using both past and future context.

Offline-Poly is learning-free and executes efficiently on CPU devices.
It incorporates category-specific strategies~\cite{li2023poly, li2024fast} for multi-class scenarios and is compatible with a diverse range of upstream trackers.
We evaluate Offline-Poly on the nuScenes~\cite{caesar2020nuscenes} and KITTI~\cite{geiger2012we} benchmarks.
On nuScenes, it achieves state-of-the-art performance with 77.6\% AMOTA.
On KITTI, it ranks first among all MOT methods with 83.00\% HOTA, significantly outperforming existing online and offline trackers.
Extensive experiments demonstrate its flexibility, strong generalization, and the effectiveness of its modular design.
We expect Offline-Poly to serve as a robust and extensible baseline for future offline 3D MOT research.
The primary contributions of this work are:


\begin{itemize}
\item We propose Offline-Poly, a robust and general framework for offline 3D MOT built upon a novel \textit{Tracking-By-Tracking} paradigm.
\item We systematically decompose offline tracking into modular components: pre-processing, matching, fusion, and refinement. 
Each module is explicitly designed to exploit resource unconstrainedness and future observability.
\item We present the \textbf{first} offline 3D MOT method supporting an arbitrary number of input tracking results, enabling flexible and effective multi-source integration.
\item Offline-Poly achieves state-of-the-art performance on nuScenes and KITTI, with 77.6\% AMOTA and 83.00\% HOTA, respectively.
Code will be released publicly.
\end{itemize}

\section{Related Work}

\subsection{Online 3D MOT}

Due to the available and accurate depth information, LiDAR is widely adopted in pioneering online 3D MOT.
Mostly LiDAR-based trackers\cite{weng20203d, pang2022simpletrack, wang2021immortal, benbarka2021score,li2023poly, li2024fast, wang2025mctrack} follow the TBD framework for its modularity, efficiency and flexibility. 
This framework typically comprises four modules: pre-processing, motion estimation, data association, and lifecycle management.
AB3DMOT~\cite{weng20203d} extends the TBD framework from 2D MOT~\cite{cao2025topic, fang2025associate} to 3D MOT, establishing a foundational baseline for subsequent works.
SimpleTrack~\cite{pang2022simpletrack} conducts a systematic analysis of the pipeline and introduces point-wise improvements.
Poly-MOT~\cite{li2023poly} reformulates multi-category tracking by accounting for inter-category object heterogeneity.
Fast-Poly~\cite{li2024fast} prioritizes runtime optimization, delivering a strong baseline that balances speed and accuracy.
TBD-based trackers are highly competitive, largely due to their modular design, which facilitates efficient algorithm analysis and rapid iteration.
Driven by rich texture cues and the emergence of query-based architectures~\cite{vaswani2017attention}, camera-based 3D MOT has advanced rapidly.
Recent methods~\cite{mutr3d, lin2023sparse4d, doll2023star, dq-track, ding2024ada, li2025hat} largely adopt the \textit{Tracking-by-Attention (TBA)} paradigm, where tracklets are represented as high-dimensional queries, and data association and tracklet updates are implicitly realized via attention mechanisms~\cite{vaswani2017attention}.
MUTR3D~\cite{mutr3d} extends DETR3D~\cite{DETR3D} to unify detection and tracking in an end-to-end manner. PF-Track~\cite{pftrack} improves trajectory consistency through past–future reasoning. DQ-Track~\cite{dq-track} and ADA-Track~\cite{ding2024ada} further enhance performance by decoupling subtasks with task-specific queries in a fully differentiable framework.

LiDAR-Camera methods~\cite{chiu2021probabilistic, wang2023interactive, lee2024dino, li2023camo, huang2021joint, wang2024multi, sadjadpour2023shastafuse, zeng2024fusiontrack} leverage the dense semantic cues from cameras and the precise spatial measurements provided by LiDAR.
Existing approaches mainly fuse multi-modal, multi-source data via result-level fusion.
Cross-modal correspondences are first established on the image~\cite{kim2021eagermot, wang2022deepfusionmot, xu2024exploiting}, followed by a motion-appearance cascade association to improve matching recall and accuracy.
The 2D and 3D features of the resulting trajectories are meticulously preserved.
Recent methods~\cite{lee2024dino, li2023camo, wang2024multi} project single-source 3D observations onto the 2D image plane to extract corresponding appearance features.
To reduce information loss, transformer-based architectures~\cite{vaswani2017attention} have been adopted~\cite{zhang2023motiontrack}.
Multi-stage attention mechanisms then implicitly construct modality-specific cost and iteratively refine the trajectory within each modality.

Although online trackers have achieved substantial progress, these onboard methods remain constrained by the latency-computation trade-off, which limits their tracking accuracy and practical suitability for 4DAL systems.
In contrast, Offline-Poly is explicitly designed for the offline setting and thoroughly optimized to exploit its unrestricted computational resources and full-scene observability.

\subsection{Offline 3D MOT}
The fundamental objective of both offline and online 3D MOT is to maintain consistent object identities across consecutive frames.
Unlike online tracking, offline 3D MOT operates without real-time constraints and has access to the entire temporal sequence, enabling global reasoning over full-scene trajectories~\cite{RoboMOT, OTOP, bitrack, PC3T}. 
The development of offline tracking has been largely motivated by 4D auto-labeling (4DAL) systems, where temporally consistent annotations are critical.
Within 4DAL pipelines~\cite{ma2023detzero, 3DAL, Auto4D, CTRL}, the offline 3D MOT module aggregates frame-wise 3D detections into 4D object trajectories.
It enables automatic generation of high-quality labels that encode both 3D spatial attributes (position, size, orientation) and 1D temporal attributes (velocity and identity).
Early 4DAL systems (3DAL~\cite{3DAL}, Auto4D~\cite{Auto4D}) directly employ online trackers~\cite{weng20203d} for offline 3D MOT. 
Subsequent methods (DetZero~\cite{ma2023detzero}, CTRL~\cite{CTRL}) enhance tracking recall by incorporating bidirectional (forward-backward) tracking and flexible trajectory termination strategies inspired by online tracking frameworks~\cite{wang2021immortal}.
Nevertheless, these systems remain fundamentally rooted in online tracking paradigms, with only limited adaptation to the offline setting.

More recently, several approaches~\cite{PC3T, wang2025mctrack, RoboMOT, bitrack, OTOP} have treated offline 3D MOT as a standalone research problem, pushing the boundaries of the field.
Early approaches primarily operate on individual tracklets.
PC3T~\cite{PC3T} employs score filtering and empty-frame interpolation to suppress ghost tracklets and enhance recall.
RethinkMOT~\cite{RoboMOT} introduces an outlier detector to remove unreliable trajectory states.
Recent methods further model relationships between trajectories within online tracking outputs.
OTOP~\cite{OTOP} leverages static map priors to construct tracklet-level association costs and completes fragmented trajectories using learning-based motion prediction.
BiTrack~\cite{bitrack} performs bidirectional tracking on a single detection set, clustering and splitting tracklets based on shared detections before merging them according to temporal continuity.
Despite these advances, most existing offline trackers remain tightly coupled to specific upstream inputs or task-specific architectural designs, limiting their generalizability.
For instance, BiTrack~\cite{bitrack} relies on observation-tracklet matching pairs inherited from online tracking, OTOP~\cite{OTOP} requires static maps for tracklet completion, and the bidirectional tracking in DetZero~\cite{ma2023detzero} is bound to the BYTETrack~\cite{zhang2022bytetrack} framework.
Moreover, key advantages unique to the offline context (multi-source integration, etc.) are still insufficiently explored.
In contrast, under the proposed \textit{Tracking-by-Tracking (TBT)} paradigm, Offline-Poly performs offline tracking solely based on upstream tracking results, without relying on frame-wise detections or intermediate association metadata.
It decouples offline optimization from specific detector or tracker architectures and systematically incorporates the intrinsic properties of the offline setting within a holistic, input-agnostic framework.

\begin{figure*}
    \centering
    \includegraphics[width=1.0\linewidth]{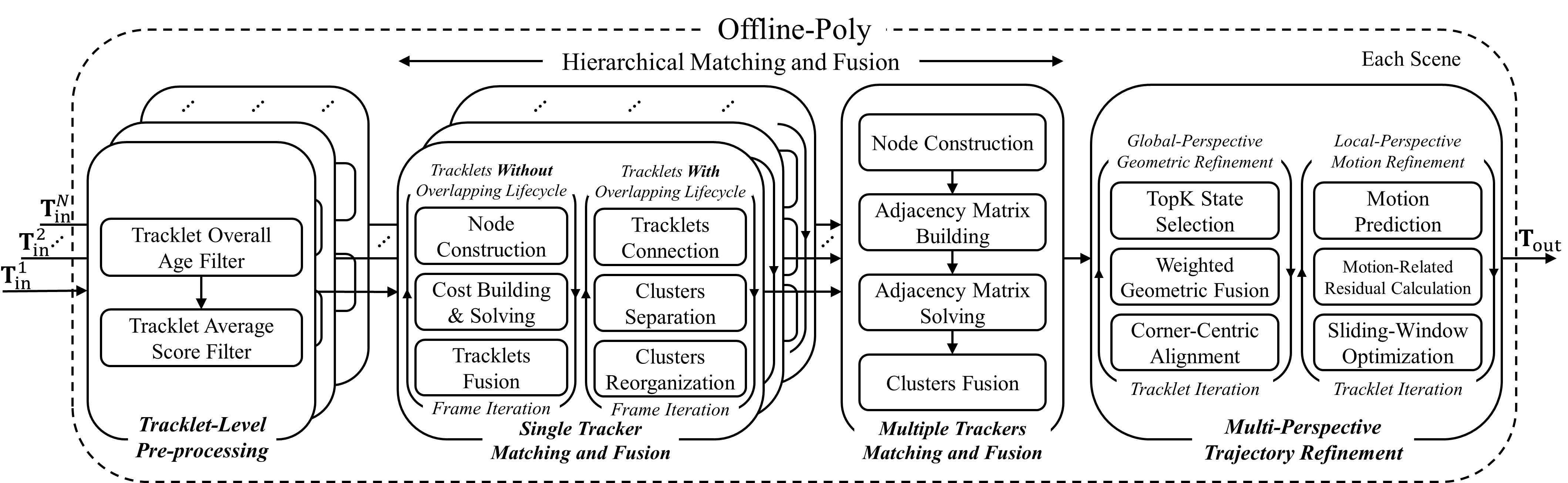}
    \caption{The pipeline of our proposed method. 
    Offline-Poly refines coarse tracking results into optimized trajectories through four stages: pre-processing, matching, fusion, and refinement. The detailed architecture is provided in~\cref{Overall_Architecture}.}
    \label{fig:overall_framework}
    \vspace{-1em}
\end{figure*}

\section{Tracking-By-Tracking}

In this section, we formalize the proposed \textit{Tracking-by-Tracking} paradigm by explicitly defining its inputs and outputs.
To ensure scalability and compatibility with diverse upstream methods, TBT adopts a tracking-centric and cascade-capable design, directly consuming finalized tracking results without requiring intermediate association metadata.

\textbf{Inputs.} 
The inputs to TBT are tracking results produced by one or multiple upstream trackers.
For a single tracker, its output is represented as a set of tracklets $\mathbf{T}_{\text{in}} = \{ T^{i}_{\text{in}} \mid i=1, 2, \dots, N^i_{\text{in}} \}$.
\(N^i_{\text{in}}\) is the number of input tracklets.
Each tracklet \(T^{i}_{\text{in}} = \{ B_{j} \mid j=1, 2, \dots, A_{i} \}\) is a sequence of states, and \(A_{i}\) denotes its age (length).
Each state \(B_{j} \in \mathbb{R}^{1 \times N_{s}}\) is a $N_{s}$-dimensional vector $(x, y, z, w, l, h, vx, vy, \theta, \mathrm{Conf}, t, \mathrm{CLS}, \mathrm{ID})$.
The vector includes the 3D center \((x, y, z)\), size \((w, l, h)\), velocity \((vx, vy)\), heading angle \(\theta\), confidence \(\mathrm{Conf}\), timestamp \(t\), object category \(\mathrm{CLS}\), and tracking identity \(\mathrm{ID}\).
Within each tracklet, \(\mathrm{ID}\) and \(\mathrm{CLS}\) remain constant, while timestamps \(t\) may be temporally discontinuous due to missed detections or occlusions.
To exploit the computational flexibility of the offline setting, TBT naturally extends to multiple input sources.
We denote the multi-source input as $\mathbb{T}_{\text{in}} = \{ \mathbf{T}^{i}_{\text{in}} \mid i=1, 2, \dots, N \}.$
\(N\) is the number of upstream tracking results.

\textbf{Output.} 
Given multi-source inputs \(\mathbb{T}_{\text{in}}\), the TBT framework produces a refined set of output tracklets:
\begin{equation}
\label{eq:offline_poly_overall}
\mathbf{T}_{\text{out}} \longleftarrow \mathrm{Tracking\text{-}By\text{-}Tracking}(\mathbb{T}_{\text{in}}),
\end{equation}
where \(\mathbf{T}_{\text{out}} = \{ T^{i}_{\text{out}} \mid i=1, 2, \dots, N_{\text{out}} \}\), and \(N_{\text{out}}\) is the number of output tracklets.

The term \textit{Tracking-by-Tracking} emphasizes that the output tracklets are constructed exclusively from input tracklets, analogous to the \textit{Tracking-By-Detection} in online MOT, where trajectories are assembled from input detections.
Notably, TBT can be extended to incorporate method-specific inputs for further performance gains. 
In this work, however, the proposed method is primarily designed to emphasize generalization and input-agnostic compatibility, as demonstrated in~\cref{table:nu_ablation}.

\section{Offline-Poly}

In this section, we describe a concrete realization of the TBT paradigm, termed Offline-Poly, detailing both the overall framework and its constituent modules.

\subsection{Overall Architecture}
\label{Overall_Architecture}

As illustrated in Fig.~\ref{fig:overall_framework}, Offline-Poly consists of four components: pre-processing, matching, fusion, and refinement.
Unless otherwise specified, all operations are performed on tracklet states represented in the global coordinate system, consistent with mainstream online trackers~\cite{wang2025mctrack, li2024fast}.
The pipeline begins with a tracklet-level pre-processing stage (\cref{pre-process}), which removes short-lived and low-quality trajectories.
Subsequently, a hierarchical matching and fusion module re-identifies and reorganizes fragmented trajectories belonging to the same object within each tracking result (\cref{single_tracker}).
Moreover, across different tracking results, inter-tracker association further consolidates multiple candidate trajectories of the same object through aggregation (\cref{Multiple_Trackers}).
Finally, a multi-perspective refinement module (\cref{refinement}) optimizes geometric and kinematic properties by leveraging both global and local trajectory information.

\vspace{-0.5em}
\subsection{Tracklet-Level Pre-Processing}
\label{pre-process}

Existing online trackers~\cite{li2023poly, pang2022simpletrack, li2024fast} employ frame-level pre-processing modules (Non-Maximum Suppression, Score Filter, etc.) to suppress false positive (FP) observations in each timestamp.
However, these frame-wise heuristics often retain high-confidence FP, which subsequently evolve into ghost tracklets. 
In offline tracking, such spurious trajectories not only degrade annotation quality but also increase the computational burden of downstream global reasoning.
Empirically, FP seldom persist consistently across consecutive frames.
Based on this observation, we adopt a tracklet-level filtering strategy that evaluates trajectory reliability holistically.
Specifically, we jointly consider tracklet age and average confidence score to identify unstable trajectories.
For each tracking result $\textbf{T}_{\text{in}} \in \mathbb{T}_{\text{in}}$, we remove any tracklet $T_{\text{in}} \in \textbf{T}_{\text{in}}$ whose age and mean confidence are both below predefined thresholds $\theta_\text{{age}}$ and $\theta_\text{{score}}$.
This criterion effectively suppresses short-lived, low-confidence ghost trajectories while preserving stable object hypotheses.
The filter yields $\mathbb{T}_{\text{pre}} = \left \{ \mathbf{T}^{i}_{\text{pre}} \mid i = 1, 2, \dots, N \right \}$.
$\mathbf{T}^{i}_{\text{pre}}$ denotes the $i$-th tracking result after filtering and is given by $\mathbf{T}^{i}_{\text{pre}} = \left \{ {T}^{j}_{\text{pre}} \mid j=1, 2, \dots, N^{i}_{\text{pre}} \right \}$.
$N^{i}_{\text{pre}}$ denotes the number of tracklets retained from the $i$-th tracker.
This design explicitly leverages future observability by assessing trajectory stability over its full temporal span rather than relying on instantaneous frame-level cues.



\begin{figure*}[t]
    \centering
    \includegraphics[width=1.0\linewidth]{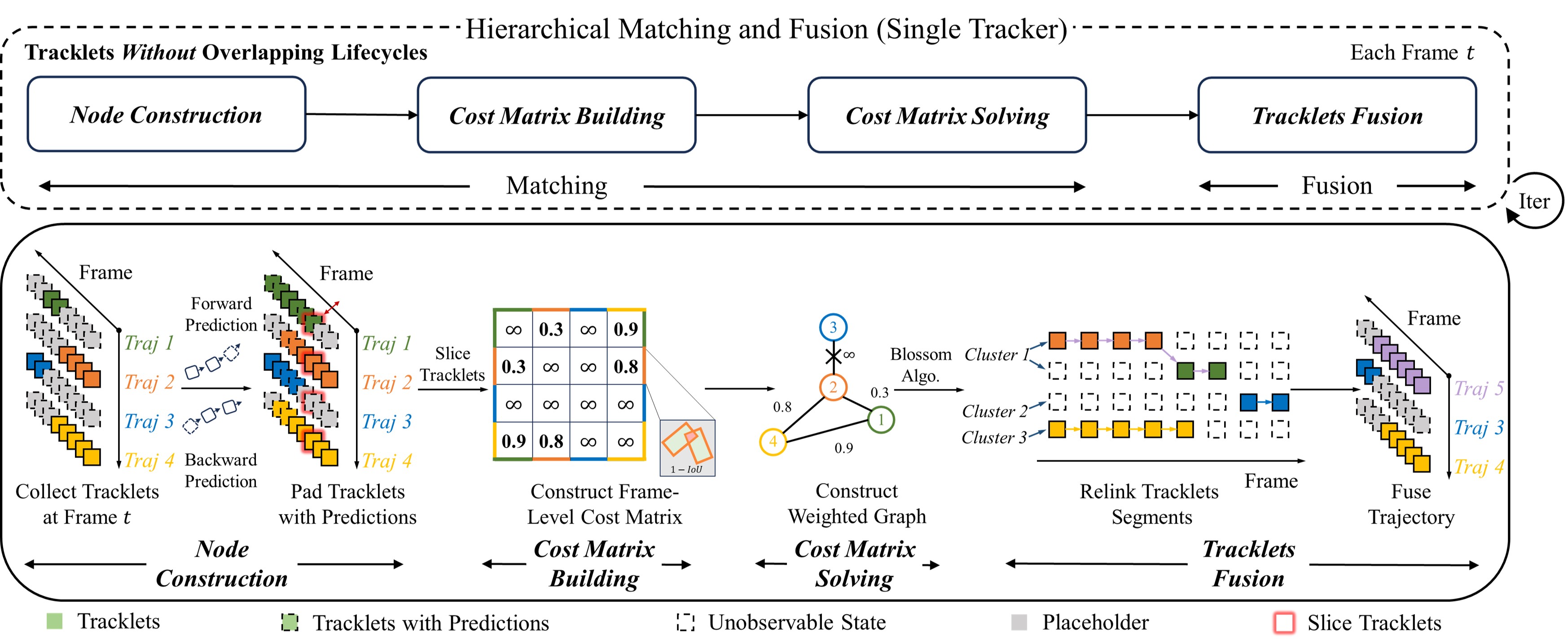}
    \caption{The pipeline of single tracker matching and fusion for tracklets without overlapping lifecycles (STWO). 
    Offline-Poly introduces a motion-based matching and fusion framework that re-identifies tracklet fragments belonging to the same object through iterative frame-by-frame processing.
    }
    \vspace{-1em}
    \label{fig:single_non_tracker}
\end{figure*}

\subsection{Hierarchical Matching and Fusion (Single Tracker)}
\label{single_tracker}

In this section, we perform re-identification and structural reorganization of individual tracking results by leveraging the future temporal context available in the offline setting. The procedure consists of two complementary stages:
\textbf{For tracklets \textit{without} overlapping lifecycles:} We introduce a novel motion-based matching framework to re-associate temporally disjoint tracklet fragments that belong to the same physical object.
\textbf{For tracklets \textit{with} overlapping lifecycles:} We disentangle tracklets that were erroneously merged from multiple objects, and restructure them into consistent trajectory segments.
These two stages are applied independently to each tracking result $\textbf{T}_{\text{pre}} \in \mathbb{T}_{\text{pre}}$.
For clarity, the procedure is illustrated using a representative tracking result $\textbf{T}_{\text{pre}}$.

\textbf{Tracklets \textit{Without} Overlapping Lifecycles (STWO).}
\label{non_overlap}
Online trackers frequently produce fragmented tracklets due to occlusion or false negative (FN) detections.
Offline tracking, benefiting from object permanence and global scene context, can mitigate these failures and improve tracking quality.
A recent offline method~\cite{OTOP} addresses this by computing and solving neural network (NN)-based affinities derived from tracking queries and map features.
In contrast, we propose an iterative, learning-free matching and fusion framework that reconnects temporally disjoint tracklets based on motion continuity.
As illustrated in Fig.~\ref{fig:single_non_tracker}, the procedure operates incrementally across frames.
Each iteration consists of four steps:
(\uppercase\expandafter{\romannumeral1}) Node generation via motion extrapolation,
(\uppercase\expandafter{\romannumeral2}) Cost matrix construction,
(\uppercase\expandafter{\romannumeral3}) Assignment solving, and
(\uppercase\expandafter{\romannumeral4}) Merging of matched tracklet fragments.
The module receives $\textbf{T}_{\text{pre}}$ as the initial input.
After each iteration, an updated set $\textbf{T}_{\text{ReID}}$ is generated and used for subsequent processing.
The following describes the procedure at frame $t$ given input set $\textbf{T}_{\text{ReID}}$.

\textit{Node Construction.}
At frame $t$, we collect the available states of all tracklets $T_{\text{ReID}} \in \textbf{T}_{\text{ReID}}$.
For tracklets without an observed state at the current frame $t$, we perform forward-backward prediction based on physical motion models~\cite{li2023poly} to estimate their states.
The prediction is formulated as:
\begin{equation}
\hat{B}_{i,t} =
\begin{cases}
f_{t_{\text{end}}\to t}(B_{i,t_{\text{end}}}),      & t > t_{\text{end}}, \\
f_{t_{\text{start}} \to t}(B_{i,t_{\text{start}}}),  & t < t_{\text{start}}, \\
\dfrac{1}{2}[f_{t_{-} \to t}(B_{i,t_{-}}) +
                       f_{t_{+} \to t}(B_{i,t_{+}})], & t_{\text{start}} < t < t_{\text{end}}.
\end{cases}
\label{eq:bi_prediction}
\end{equation}
where $B_{i, t}$ denotes the observed state and $\hat{B}_{i,t}$ denotes the predicted 3D state.
The propagation function $f_{t_0 \to t}(B_{i,t_0})$ evolves a state from frame $t_o$ to $t$ according to an explicit motion model~\cite{li2023poly} (Constant Velocity, Constant Turn Rate and Acceleration, etc.).
$t_{\text{start}}$ and $t_{\text{end}}$ are the first and last frames in which the tracklet is observed.
$t_-$ and $t_+$ are the nearest observed frames before and after frame $t$.
Predictions exceeding 1 second are discarded to prevent drift.
Finally, observed and valid predicted states are aggregated into a unified node set $\mathbf{B}_{\text{pred}} \in \mathbb{R}^{N_{\text{ReID}} \times N_{s}}$.
$\mathbf{B}_{\text{pred}}$ is used for subsequent cost computation, where $N_{\text{ReID}}$ is the number of tracklets in $\textbf{T}_{\text{ReID}}$.

\textit{Cost Matrix Building.}
Given $\mathbf{B}_{\text{pred}}$, we compute a symmetric cost matrix $C_{\text{ReID}, t} \in \mathbb{R}^{N_{\text{ReID}} \times N_{\text{ReID}}}$ to quantify pairwise dissimilarities between tracklets at frame $t$.
We adopt the \textit{Intersection over Union (IoU)} as the affinity metric, and define the cost as:
\begin{equation}
\label{single:frame_level}
    C_{\text{ReID}, t}^{i, j} = 1 - \frac{\Lambda(\tilde{B}_{i, t} \cap \tilde{B}_{j, t})}{\Lambda(\tilde{B}_{i, t} \cup \tilde{B}_{j, t})}, 
\end{equation}
where $C_{\text{ReID}, t}^{i,j}$ denotes the cost between tracklets $T^{i}_{\text{ReID}}$ and $T^{j}_{\text{ReID}}$ at frame $t$.
$\tilde{B}_{i, t} \in \mathbf{B}_{\text{pred}}$ represents the observed or 
motion-predicted state of $T^{i}_{\text{ReID}}$ at frame $t$.
Self-assignment is prevented by setting the diagonal entries of $C^{i, j}_{\text{ReID}, t}$ to infinity, and costs involving invalid states are likewise set to infinity.
Here, $\tilde{B}_{i, t} \cap \tilde{B}_{j, t}$ and 
$\tilde{B}_{i, t} \cup \tilde{B}_{j, t}$ denote the geometric intersection 
and union, respectively. 
The operator $\Lambda(\cdot)$ computes the area in the Bird’s-Eye-View (BEV) space 
or the volume in the 3D space.

\textit{Solving the Matrix.}
Assuming a one-to-one correspondence between trajectory fragments, we employ a graph-based assignment algorithm to identify optimal matching pairs based on $C_{\text{ReID}, t}$.
Specifically, we adopt the blossom algorithm~\cite{edmonds1965paths} to solve the assignment problem, which efficiently computes globally optimal pairings among tracklet nodes under a unified cost objective.
To ensure association reliability, only matches with costs lower than a predefined threshold $\theta_{\text{blo}}$ are retained.
After assignment, the tracklets are partitioned into matched pairs $\mathbb{T}^{\text{m}}_{\text{ReID}}$ and unmatched trajectories $\textbf{T}^{\text{um}}_{\text{ReID}} \in \textbf{T}_{\text{ReID}}$.
The set of matched pairs is defined as:
\begin{equation}
\label{single:matching_pair}
\mathbb{T}^{\text{m}}_{\text{ReID}} = \left\{ \left( T^{i}_{\text{ReID}}, T^{j}_{\text{ReID}} \right) \;\middle|\; i, j \in \{1, \dots, N_{\text{ReID}}\},\ i \ne j \right\}.
\end{equation}



\textit{Fusion and Iteration.}
Each element in $\mathbb{T}^{\text{m}}_{\text{ReID}}$ 
corresponds to a pair of tracklet fragments originating from the same physical object.
Each pair is fused into a single trajectory and assigned a new tracking ID. 
The motion model prediction (\cref{eq:bi_prediction}) is then used to interpolate missing observations within the merged tracklet.
After processing all pairs, $\mathbb{T}^{\text{m}}_{\text{ReID}}$ is consolidated into the merged tracklet set $\mathbf{T}^{\text{m}}_{\text{ReID}}$.
By integrating $\mathbf{T}^{\text{m}}_{\text{ReID}}$ with the unmatched set $\mathbf{T}^{\text{um}}_{\text{ReID}}$, we obtain the updated tracklet set $\mathbf{T}_{\text{ReID}}$.
Since a single object may be fragmented multiple times, the matching-fusion procedure is applied iteratively to progressively refine and consolidate trajectories, enabled by the absence of real-time constraints in the offline setting.

\begin{figure*}
    \centering
    \includegraphics[width=1\linewidth]{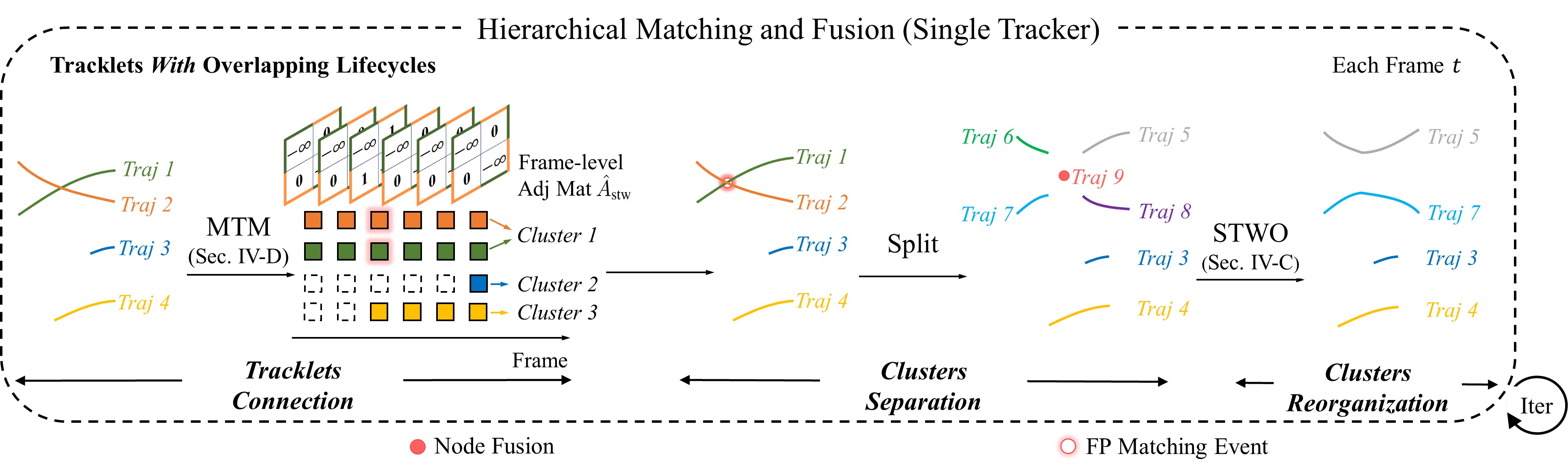}
    \caption{The pipeline of single matching and fusion for tracklets with overlapping lifecycles (STW).
    It matches tracklets with high geometric similarity, then reorganizes and disentangles the distinct objects embedded within a single tracklet.
    We present the process for the $i$-th cluster.}
    \vspace{-1em}
    \label{fig:single_overlap_tracker}
\end{figure*}

\textbf{Tracklets \textit{With} Overlapping Lifecycles (STW).}
Online trackers also suffer from FP associations, in which similar observations are incorrectly merged, and distinct objects are fused into a single tracklet.
Departing from frame-wise and learning-based anomaly analysis~\cite{RoboMOT}, we introduce a tracklet-centric, learning-free matching and restructuring framework that connects, disentangles, and restructures intertwined trajectories.
This framework comprises three key stages:
(\uppercase\expandafter{\romannumeral1}) Matching similar tracklets,
(\uppercase\expandafter{\romannumeral2}) Separating tracklet clusters, and
(\uppercase\expandafter{\romannumeral3}) Reorganizing tracklets.
The module receives $\textbf{T}_{\text{ReID}}$ as the initial input (the operation order between STWO and STW is ablated in~\cref{table:nu_ablation}).

\textit{Tracklets Connection.}
We first match tracklets in $\textbf{T}_{\text{ReID}}$ that exhibit high geometric similarity, as this resemblance can lead to FP associations in geometry-driven online trackers~\cite{li2023poly}.
To capture and solve these correlations, we reuse the multi-tracker matching module described in \cref{Multiple_Trackers}.
A frame-level adjacency matrix $\hat{A}_{\text{stw}} \in \mathbb{R}^{L_s \times N_{\text{ReID}} \times N_{\text{ReID}}}$ is constructed using \textit{generalized IoU (gIoU)} and \cref{eq:binarize}, where $L_s$ denotes the scene length.
Applying the search algorithm in~\cref{multi:soving}, geometrically similar tracklets are grouped into clusters $\mathbb{T}_{\text{cluster}}$.
Each cluster is represented as $\mathbf{T}_{\text{cluster}} = \left\{ T_{\text{ReID}}^i \mid i = 1, \dots, N_{\text{cluster}} \right\}$.
$N_{\text{cluster}}$ is the number of tracklets.
Details of the clustering procedure are provided in~\cref{Multiple_Trackers}.
We then perform separation and reorganization on each cluster.
For clarity, we illustrate the subsequent disambiguation process using a representative cluster $\mathbf{T}_{\text{cluster}}$.


\textit{Clusters Separation.}
FP associations typically corrupt trajectories locally around the mismatching event, while long-term segments before and after remain reliable.
Leveraging this property, \(\mathbf{T}_{\text{cluster}}\) is decomposed into reliable trajectory segments and ambiguous entangled nodes.
Specifically, at each frame \(t\), we extract connected components of the adjacency graph \(\hat{A}_{\text{stw}, t}\in\mathbb{R}^{N_{\text{cluster}}\times N_{\text{cluster}}}\).
Tracklets that are directly or indirectly connected within a frame are marked as entangled.
Each involved tracklet is then split at the connected frames into pre- and post-event segments, which are preserved as reliable trajectory segments.
FP matching can persist across many frames, causing a single tracklet to be segmented multiple times.
All entangled components are merged into new tracklets initialized with age 1 (fusion details in~\cref{Multiple_Trackers}).
Finally, the reliable segments and newly formed entangled-node tracklets are aggregated to produce updated clusters \(\mathbf{T}_{\text{cluster}}\) with non-overlapping lifecycles.

\textit{Clusters Reorganization.}
Leveraging full-sequence visibility, we associate the resulting segments using bidirectional motion prediction to reconstruct complete trajectories.
Given the updated $\mathbf{T}_{\text{cluster}}$, we apply the STWO module (\cref{non_overlap}) to iteratively relink tracklets with high cross-frame similarity.

\textbf{Output.}
Through the above re-identification and reorganization processes, each preliminary tracking result $\mathbf{T}_{\text{pre}}$ is refined into $\mathbf{T}_{\text{sin}}$.
The outputs from all trackers, $\mathbb{T}_{\text{sin}} = \{\mathbf{T}_{\text{sin}}^{i}, \mid i = 1, \dots, N\}$, are subsequently merged via multiple trackers matching and fusion, exploiting cross-source complementarity to produce more complete and consistent object trajectories.

\begin{figure*}
    \centering
    \includegraphics[width=1\linewidth]{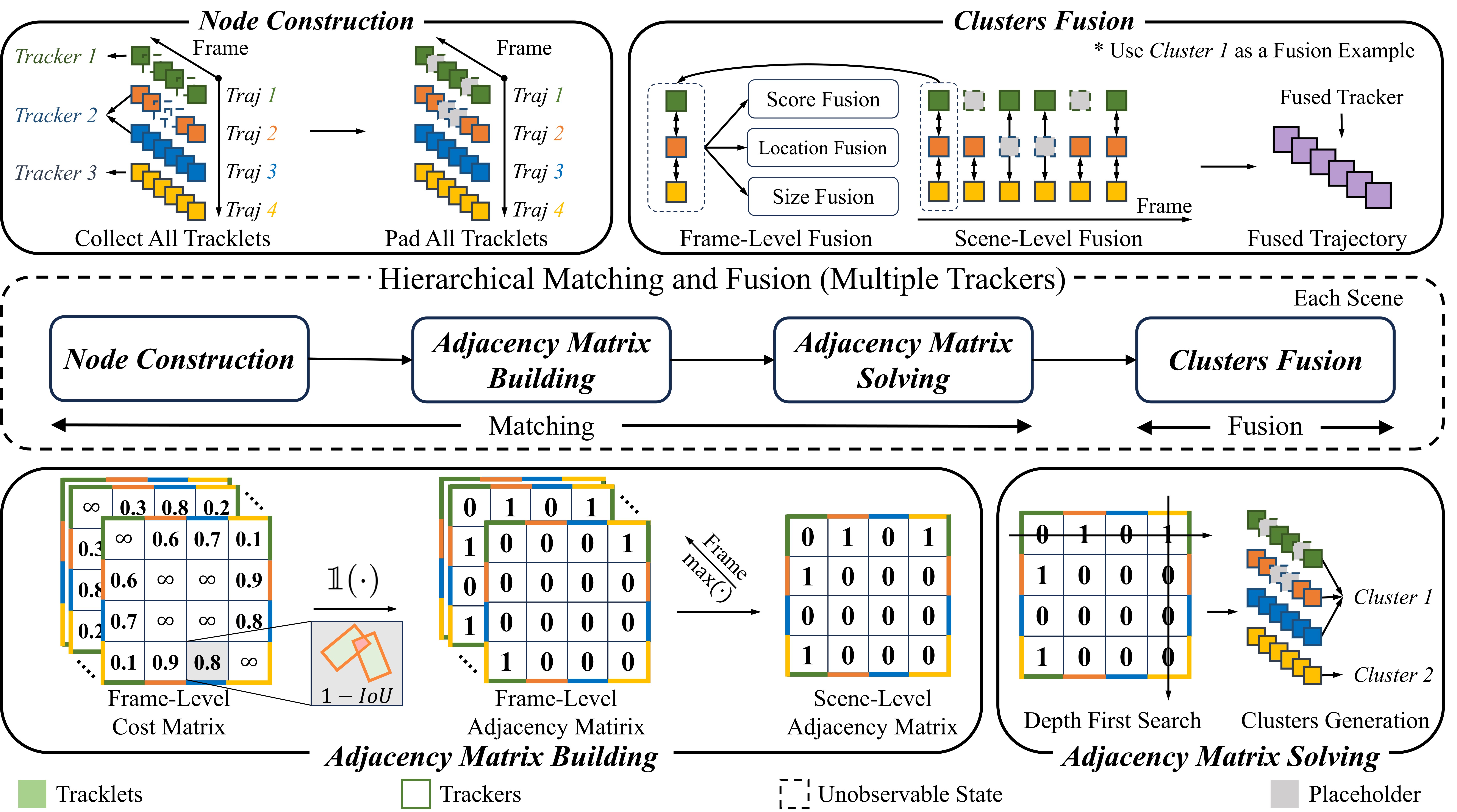}
    \caption{The pipeline of hierarchical matching and fusion (multiple trackers). Offline-Poly integrates cross-tracker observations of the same object by constructing scene-level tracklet similarity, producing more complete and consistent tracklet representations.}
    \vspace{-1em}
    \label{fig:mutli_tracker}
\end{figure*}

\subsection{Hierarchical Matching and Fusion (Multiple Trackers)}
\label{Multiple_Trackers}

In offline 3D MOT, computational and latency constraints are largely relaxed.
However, most existing approaches rely on a single source~\cite{OTOP, RoboMOT, bitrack}, neglecting the opportunity to integrate multiple tracking outputs.
Different trackers can capture complementary aspects of the same object. Aggregating these heterogeneous hypotheses produces more complete and accurate trajectories.
Accordingly, as illustrated in Fig.~\ref{fig:mutli_tracker}, we propose a multi-tracker matching and fusion module to consolidate tracking results from multiple sources.

\textbf{Matching.}
To identify candidate tracklets corresponding to the same object, we first aggregate tracklets from all trackers and compute their pairwise affinities.
Unlike prior approaches that depend on specific input features (static maps~\cite{OTOP} or online observations~\cite{bitrack}, etc.), our framework adopts a geometry-based and lightweight design, enabling efficient association across heterogeneous tracking outputs and compatibility with diverse affinity metrics.
As shown in Fig.~\ref{fig:mutli_tracker}, the proposed matching process consists of three stages:
(\uppercase\expandafter{\romannumeral1}) Node construction from all input trajectories, (\uppercase\expandafter{\romannumeral2}) Adjacency matrix building, and 
(\uppercase\expandafter{\romannumeral3}) Adjacency matrix solving.

\textit{Node Construction.} 
We begin by concatenating trajectories from all tracking results $\mathbb{T}_{\text{sin}}$:
\begin{equation}
\label{eq:edge_construction}
    \mathbf{T}_{\text{pad}} = \left\{ T^{j}_{\text{pad}} \mid j = 1, 2, \dots, N_{\text{all}} \right\}, \quad N_{\text{all}} = \sum_{i=1}^{N} N^{i}_{\text{sin}},
\end{equation}
where $N^{i}_{\text{sin}}$ is the number of tracklets of $i$-th single tracking results $\mathbf{T}^{i}_{\text{sin}}$.
To enable efficient parallel computation, each trajectory is padded to the full scene length $L_s$ using placeholder values (\textit{NaN}), accompanied by a binary mask indicating valid entries.
The resulting tracklets, $\mathbf{T}_{\text{pad}} \in \mathbb{R}^{N_{\text{all}} \times L_s \times N_{s}}$, serve as node representations for constructing the adjacency matrix.

\textit{Adjacency Matrix Building.}
In contrast to the one-to-one assumption commonly used in online 3D MOT~\cite{li2023poly, li2024fast} and other offline methods~\cite{OTOP, RoboMOT}, we adopt a one-to-many formulation, allowing multiple trajectory hypotheses from distinct trackers to correspond to the same object.
We formulate the candidate trajectory identification problem as constructing an adjacency matrix among tracklets.
Tracklets belonging to the same object are expected to exhibit strong geometric or appearance consistency across frames, \textit{i.e.}, spatio-temporal coherence.
Guided by this principle, as shown in Fig. \ref{fig:mutli_tracker}, we build the adjacency matrix based on $\mathbf{T}_{\text{pad}}$.

For each frame $t$, a cost matrix $C_{\text{multi}, t} \in \mathbb{R}^{N_{\text{all}} \times N_{\text{all}}}$ is computed to quantify spatial affinities between tracklet pairs.
This matrix is derived from the tracklet states at frame $t$ using a geometry-based metric (\textit{IoU}), consistent with~\cref{single:frame_level}.
The matrix is binarized as:
\begin{equation}
\label{eq:binarize}
A_{\text{multi}, t}^{i, j} = \mathbbm{1}\left(C_{\text{multi}, t}^{i, j} < \theta_{\text{multi}} \right),
\end{equation}
where $\mathbbm{1}(\cdot)$ denotes the indicator function, and $\theta_{\text{multi}}$ is a fixed connectivity threshold (1 for connected, 0 for disconnected).
To incorporate full-sequence information, frame-level adjacency matrices are aggregated temporally:
\begin{equation}
\label{multi:frame_level}
    \hat{A}_{\text{multi}} = \text{Concat}(A_{\text{multi}, 1},\dots,A_{\text{multi}, L_{s}}), 
\end{equation}
where $\hat{A}_{\text{multi}} \in \mathbb{R}^{L_{s} \times N_{\text{all}} \times N_{\text{all}}}$ is the aggregated frame-level adjacency matrix.
$\text{Concat}(\cdot)$ denotes the concatenate function.
Finally, we compress $\hat{A}_{\text{multi}}$ along the temporal axis to determine scene-level connectivity:
\begin{equation}
\label{eq:squeeze}
\tilde{A}^{i,j}_{\text{multi}} = \max_{t=1}^{L_s} \hat{A}^{i,j}_{\text{multi}}[t, i, j],
\end{equation}
where $\tilde{A}_{\text{multi}} \in \mathbb{R}^{N_{\text{all}} \times N_{\text{all}}} $ represents the scene-level adjacency matrix over $\mathbf{T}_{\text{pad}}$.
This lightweight construction naturally generalizes to heterogeneous tracking outputs and supports various similarity metrics.
In~\cref{eq:squeeze}, max pooling is employed to achieve high recall with minimal computational overhead, despite its susceptibility to spurious edge responses. 
Nevertheless, as shown in Table~\ref{table:nu_ablation}, the MTM module demonstrates strong effectiveness and robustness. 
Developing more reliable temporal aggregation strategies remains future work.

\textit{Adjacency Matrix Solving.}
We perform a search algorithm on $\tilde{A}_{\text{multi}}$ to identify trajectories with direct or indirect adjacency relationships.
The process is formulated as:
\begin{equation}
\label{multi:soving}
\mathbb{T}_{\text{multi}} \longleftarrow \text{DFS}(\tilde{A}_{\text{multi}}),
\end{equation}
where $\text{DFS}(\cdot)$ is the depth-first search algorithm.
$\mathbb{T}_{\text{multi}}$ is the resulting set of tracklet clusters.
Each cluster $\mathbf{T}_{\text{cluster}} = \left\{ T^{i}_{\text{cluster}} \mid i = 1, \dots, N_{\text{cluster}} \right\}$ aggregates observations of the same object across diverse tracking outputs.
$N_{\text{cluster}}$ varies by instance, reflecting differences in object observability.

\textbf{Fusion.}
Finally, each trajectory cluster $\mathbf{T}_{\text{cluster}}^{i}$ is merged using a simple heuristic.
Fusion is attribute-agnostic: per-frame attributes are computed as weighted averages of all valid observations.
The fused state, integrating complementary information from multiple trackers, is inserted into the merged tracklet.
Category-consistent association is enforced to maintain semantic consistency~\cite{li2023poly,li2024fast}.
A unique tracking ID is then assigned to each fused trajectory, yielding the final set $\mathbf{T}_{\text{multi}} = \left \{ T^{i}_{\text{multi}} \mid i = 1, \dots, N_{\text{out}} \right \}$, where $N_{\text{out}}$ denotes the number of trajectories.
The resulting trajectories are further refined using temporal information.








\begin{table}[t!]
\centering
\caption{A comparison of Offline-Poly across distinct detectors and upstream trackers on the nuScenes val set. 
CP: CenterPoint~\cite{yin2021center}. 
FP: Fast-Poly~\cite{li2024fast}. 
MV: MV2DFusion~\cite{mv2dfusion}. 
ADA: ADA-Track~\cite{ding2024ada}.
CR: Cascade R-CNN~\cite{cai2018cascade}.
OP: Offline-Poly.
-F: Forward tracking. 
-B: Backward tracking.}
\label{table:scale_nusc}
\renewcommand{\arraystretch}{1}
\setlength{\tabcolsep}{1.2mm}{
\begin{tabular}{ccccccc}
\toprule
\textbf{Index} & \textbf{Detector(s)} & \textbf{Tracker(s)} & \textbf{OP} & \textbf{AMOTA}$\uparrow$ & \textbf{MOTA}$\uparrow$ \\ \midrule
- & CP\&CR  & EagerMOT~\cite{kim2021eagermot} & $\times$ & 71.2      & -          \\
-        & CP  & Poly-MOT\cite{li2023poly} & $\times$ & 73.1      & 61.9        \\
- & CP  & MCTrack\cite{wang2025mctrack} & $\times$ & 74.0      & 64.0      \\	
\midrule
EXP1         & CP        & FP-F       & $\times$      & 73.7   & 63.2        \\
\rowcolor{gray!25} EXP2         & CP        & FP-F       & \checkmark      & 74.8 \textbf{{\scriptsize (+1.1)}}  & 65.0 \textbf{{\scriptsize (+1.8)}}       \\
EXP3         & CP        & FP-B       & $\times$      & 74.3   & 63.8        \\
\rowcolor{gray!25} EXP4         & CP        & FP-B       & \checkmark      & 75.5 \textbf{{\scriptsize (+1.2)}} & 64.8 \textbf{{\scriptsize (+1.0)}}   \\
\rowcolor{gray!25} EXP5         & CP        & FP-F\&FP-B      & \checkmark       & 76.2 \textbf{{\scriptsize (+2.5)}} & 64.8 \textbf{{\scriptsize (+1.6)}}  \\
\midrule
EXP6         & MV         & FP-F       & $\times$      & 79.8   & 69.3        \\
\rowcolor{gray!25} EXP7         & MV        & FP-F       & \checkmark      & 80.5 \textbf{{\scriptsize (+0.7)}} & 70.2 \textbf{{\scriptsize (+0.9)}}\\
EXP8         & MV        & FP-B       & $\times$      & 79.9   & 69.4   \\
\rowcolor{gray!25} EXP9         & MV        & FP-B       & \checkmark      & 80.8 \textbf{{\scriptsize (+0.9)}}  & 70.5 \textbf{{\scriptsize (+1.1)}}    \\
EXP10         & ADA        & ADA       & $\times$      & 38.4 & 34.7  \\
\rowcolor{gray!25} EXP11         & ADA        & ADA       & \checkmark      & 41.1 \textbf{{\scriptsize(+2.7)}} & 37.6 \textbf{{\scriptsize (+2.9)}} \\
\midrule
\rowcolor{gray!25} EXP12         & MV        & FP-F\&FP-B      & \checkmark   & 81.3 \textbf{{\scriptsize (+1.5)}} & 70.6 \textbf{{\scriptsize (+1.3)}} \\
\rowcolor{gray!25} EXP13 & MV\&ADA & FP-F\&FP-B\&ADA & \checkmark & 81.4 \textbf{{\scriptsize (+1.6)}} & 70.7 \textbf{{\scriptsize (+1.4)}}\\

\bottomrule
\end{tabular}}
\end{table}

\begin{table}[t]
\centering
\caption{A comparison between Offline-Poly with other methods on the KITTI test set. 
$\dagger$: We re-implement Fast-Poly~\cite{li2024fast} on KITTI. 
VC: VirConv~\cite{VirConv}.
PG: PointGNN~\cite{shi2020point}.
PR: PointRCNN~\cite{shi2019pointrcnn}.
PT: PointTrack~\cite{xu2021segment}.
-F: Forward tracking. 
-B: Backward tracking.
Offline: Offline 3D MOT.}
\label{table:comparison_kitti}
\renewcommand{\arraystretch}{1}
\setlength{\tabcolsep}{1.3mm}{
\begin{tabular}{ccccc}
\toprule
\textbf{Detector} & \textbf{Tracker} & \textbf{Offline} & \textbf{HOTA}$\uparrow$ & \textbf{MOTA}$\uparrow$\\ \midrule
PG    & CAMO-MOT~\cite{li2023camo}  & $\times$      & 79.95   & 90.38         \\
PR       & RethinkMOT~\cite{RoboMOT} & \checkmark  & 80.39  & 91.53         \\
VC    & PC3T~\cite{wu20213d}       & $\times$      & 81.87   & 90.24         \\
VC  & MCTrack~\cite{wang2025mctrack}       & $\times$   & 81.07 & 89.81      \\
VC  & MCTrack~\cite{wang2025mctrack}       & \checkmark    & 82.56 & 91.62      \\
VC\&PT         & BiTrack~\cite{bitrack}       & $\times$    & 81.59 & 91.07      \\
VC\&PT         & BiTrack~\cite{bitrack}       & \checkmark    & 82.69 & 91.65      \\
\cmidrule(lr){1-5}
VC & Fast-Poly-F~\cite{li2024fast}$\dagger$ & $\times$ & 81.15 & 89.93 \\
VC & Fast-Poly-B~\cite{li2024fast}$\dagger$ & \checkmark & 81.58 & 90.70 \\
\rowcolor{gray!25} VC & \textbf{Offline-Poly (Ours)} & \checkmark & \textbf{83.00} & \textbf{93.19} \\
\bottomrule
\end{tabular}}
\vspace{-1.5em}
\end{table}

\subsection{Multi-Perspective Trajectory Refinement}
\label{refinement}

The fusion mechanism outlined above operates primarily at the frame level, emphasizing spatial consistency across multiple trackers.
We further enhance trajectory quality in the temporal domain by incorporating motion cues. 
Trajectory refinement consists of two stages:
(\uppercase\expandafter{\romannumeral1}) Correcting geometric attributes using global trajectory information to provide more accurate priors for motion estimation,
(\uppercase\expandafter{\romannumeral2}) Optimizing motion-related attributes by integrating past-future observations through explicit motion models.

\textbf{Global-Perspective.}
A fundamental property of rigid objects is the temporal invariance of physical size.
For rigid objects in $\mathbf{T}_{\text{multi}}$, the Top-$K$ most reliable observations for each tracklet are selected according to per-frame tracking scores, where $K$ is manually specified.
The optimized 3D size is then computed as a confidence-weighted average, with weights obtained via softmax normalization of the tracking scores.
This TopK selection mitigates the impact of low-quality observations on size regression.
With the refined 3D size, we further adjust the 3D center using the corner-centric alignment strategy proposed in~\cite{Auto4D}.
This design is motivated by the observation that the L-shaped inflection point~\cite{Lshape} provides a more reliable geometric reference than the inferred box center.
The globally refined trajectories are denoted as $\mathbf{T}_{\text{glo}}$.

\textbf{Local-Perspective.}
Online trackers typically utilize Markov filters (Kalman Filter~\cite{kalman1960new}, etc.) for motion estimation due to real-time constraints.
However, within short temporal windows, object motion exhibits inter-frame correlations, enabling improved estimation by leveraging both past and future observations.
Recent methods refine tracklets using 3D NN-based detectors~\cite{ma2023detzero, lin2023sparse4d, 3DAL, CTRL} that implicitly encode temporal cues from raw sensor data.
In contrast, we propose an unsupervised and interpretable sliding-window optimization that operates solely on structured bounding box representations, ensuring a lightweight design and strong generalization.
Each tracklet is refined independently within a fixed-length temporal window.

For clarity, consider a representative trajectory $T_{\text{glo}}$.
For its state $B_{t}$ at frame $t$, observations from $M$ neighboring frames form a local temporal window.
Sliding-window optimization is performed by minimizing Mean Squared Error (MSE) between the predicted state and observed states within this window.
Only motion-related attributes are optimized, including the 3D center, velocity, and heading angle. 
Other state components remain fixed.
For simplicity, the optimization variable is still denoted as $B_t$.
The optimization objective is formulated as:
\begin{equation}
\label{refine:local_ops}
    \min \left (\sum_{\Delta t =-M/2}^{M/2} (f_{t \to t + \Delta t}(B_{t}) - B_{t+\Delta t})^2   \right ) ,
\end{equation}
where $f_{t \to t + \Delta t}(B_{t})$ is the state transition function defined in~\cref{eq:bi_prediction}.
The optimization is solved using the Levenberg-Marquardt algorithm~\cite{LM_ops}, which combines local curvature information with gradient-descent updates.
This sliding-window optimization captures short-term temporal dynamics while enforcing physically plausible motion through explicit kinematic constraints.
After refining all trajectories, the final set is denoted as $\mathbf{T}_{\text{out}} = \left \{ {T}^{i}_{\text{out}} \mid i=1, 2, \dots, N_{\text{out}} \right \}$, which constitutes the output of Offline-Poly.

\begin{table*}[t!]
        \centering
        \caption{The ablation studies of each module on the nuScenes val set.
        \textbf{Age}: Age Filter.
        \textbf{Score}: Score Filter.
        \textbf{STWO}: Single Tracker WithOut Overlapping Lifecycle.
        \textbf{STW}: Single Tracker With Overlapping Lifecycle.
        \textbf{MT}: Multiple Trackers.
        \textbf{GO}: Global-Perspective.
        \textbf{LO}: Local-Perspective.
        $\rightarrow$: Operation order.
        Lines 1-9 use forward Fast-Poly with CenterPoint as the upstream tracker, while Lines 10-13 use its bidirectional variant.
        }
        \label{table:nu_ablation}
        \renewcommand{\arraystretch}{1}
        \setlength{\tabcolsep}{1.6mm}{
        \begin{tabular}{cccccccc|ccc|ccc}
        \toprule
        \multirow{2.5}{*}{\textbf{Index}} & \multicolumn{2}{c}{\textbf{Pre-processing}} & \multicolumn{3}{c}{\textbf{Matching and Fusion}} & \multicolumn{2}{c}{\textbf{Refinement}}  & \multicolumn{3}{c}{\textbf{Primarily Metric}} & \multicolumn{3}{c}{\textbf{Secondary Metric}} \\ 
        \cmidrule(lr){2-3} \cmidrule(lr){4-6} \cmidrule(lr){7-8} \cmidrule(lr){9-11} \cmidrule(lr){12-14}
                               & \textbf{Age} & \textbf{Score}   & \textbf{STWO}        & \textbf{STW}        & \textbf{MT}        & \textbf{GO} & \textbf{LO} &\textbf{AMOTA}$\uparrow$ & \textbf{MOTA}$\uparrow$ & \textbf{AMOTP}$\downarrow$ &\textbf{IDS}$\downarrow$ & \textbf{FP}$\downarrow$ & \textbf{FN}$\downarrow$  \\ \midrule
        \textbf{Baseline}       & \textbf{\text{--}}        & \textbf{\text{--}}          & \textbf{\text{--}}          & \textbf{\text{--}}          & \textbf{\text{--}}          & \textbf{\text{--}}    & \textbf{\text{--}}             & 73.7                      & 63.2                      & 0.530                    & 414                     & 14713                    & 15900                \\
        \textbf{EXP1}       & \textbf{\text{--}}        & \checkmark          & \textbf{\text{--}}          & \textbf{\text{--}}          & \textbf{\text{--}}          & \textbf{\text{--}}    & \textbf{\text{--}}             & 69.9                      & 62.2                      & 0.678                    & 404                     & 14272                    & 16473                \\
        \textbf{EXP2}       & \checkmark        & \textbf{\text{--}}          & \textbf{\text{--}}          & \textbf{\text{--}}          & \textbf{\text{--}}          & \textbf{\text{--}}    & \textbf{\text{--}}             & 72.0                      & 62.7                      & 0.584                     & 258                     & 13982                    & 16995                \\
        \textbf{EXP3}       & \checkmark             & \checkmark           & \textbf{\text{--}}          & \textbf{\text{--}}          & \textbf{\text{--}}          & \textbf{\text{--}}  & \textbf{\text{--}}               & 74.0                      & 63.4                      & 0.536                     & 408                     & 14640                    & 15955                \\
        
        \textbf{EXP4}        & \checkmark            & \checkmark          & \textbf{\text{--}}          & \textbf{\text{--}}          & \textbf{\text{--}}          & \checkmark  & \checkmark   (\textbf{LO} $\rightarrow$ \textbf{GO})            & 74.5                      & 64.3                      & 0.516                     & 391                     & 14243                    & 15924                \\
        \textbf{EXP5}        & \checkmark            & \checkmark          & \textbf{\text{--}}          & \textbf{\text{--}}          & \textbf{\text{--}}          & \checkmark  & \checkmark                & 74.6                      & 64.7                      & 0.514                     & 387                     & 14219                    & 15882                 \\
        
        \textbf{EXP6}        & \checkmark            & \checkmark          & \checkmark          & \textbf{\text{--}}          & \textbf{\text{--}}          & \checkmark  & \checkmark               & 74.7                      & 64.7                      & 0.513                     & 372                     & 14300                    & 15902                \\
        \textbf{EXP7}       & \checkmark             & \checkmark          & \checkmark          & \checkmark (\textbf{STW} $\rightarrow$ \textbf{STWO})          & \textbf{\text{--}}          & \checkmark  & \checkmark         & 74.7                     & 64.9                      & 0.512                     & 353                     & 14019                   & 16135                \\
        \textbf{EXP8}       & \checkmark             & \checkmark          & \checkmark          & \checkmark          & \textbf{\text{--}}          & \checkmark  & \checkmark        & 74.8                      & 65.0                      & 0.518                     & 373                     & 13691                    & 16320                \\

        \midrule

        \textbf{EXP9}      & \checkmark              & \checkmark          & \checkmark          & \checkmark           & \checkmark          & \textbf{\text{--}}   & \textbf{\text{--}}              & 74.8                      & 63.6                      & 0.522                     & 266                     & 12479                    & 17598                \\
        \textbf{EXP10}       & \checkmark             & \checkmark           & \checkmark          & \checkmark          & \checkmark          & \checkmark        & \textbf{\text{--}}         & 75.3                      & 64.0                      & 0.515                     & 272                     & 12381                    & 17604                \\
        
        \textbf{EXP11}       & \checkmark             & \checkmark           & \checkmark          & \checkmark          & \checkmark          & \checkmark        & \checkmark (\textbf{LO} $\rightarrow$ \textbf{GO})        & 75.9                      & 64.7                      & 0.507                     & 251                     & 12149                    & 17634                \\
        \textbf{EXP12}       & \checkmark             & \checkmark           & \checkmark          & \checkmark          & \checkmark          & \checkmark        & \checkmark          & 76.2                      & 64.8                      & 0.504                     & 251                     & 12114                    & 17630                \\
        \bottomrule
        \end{tabular}}
        \vspace{-1.5em}
        \end{table*}

\section{Experiments}

\subsection{Datasets and Evaluation Metrics}
\textbf{nuScenes~\cite{caesar2020nuscenes}} is a large-scale autonomous driving benchmark comprising 1 LiDAR, 6 cameras, and 5 RaDARs. 
It consists of 700 training, 150 validation, and 150 test scenes, each lasting 20 seconds and covering diverse driving scenarios. 
Tracking annotations are provided at 2 Hz for 7 categories: \textit{Car}, \textit{Bicycle (Bic)}, \textit{Motorcycle (Moto)}, \textit{Pedestrian (Ped)}, \textit{Bus}, \textit{Trailer (Tra)}, and \textit{Truck (Tru)}.
The primary tracking metrics are Average MOT Accuracy (AMOTA), MOTA, and Average MOT Precision (AMOTP).
IDS, FP, and FN are additionally reported as secondary indicators.
MOTP, IDS, FP, and FN are reported for the configuration that achieves the highest MOTA.
We also report the detection metric ASE (Average Scale Error) to quantitatively assess the effect of global refinement.

\textbf{KITTI~\cite{geiger2012we}} comprises 21 training and 29 test sequences.
Data are collected using a LiDAR and stereo cameras at 10 Hz.
The benchmark evaluates long-term tracking of two categories: \textit{Car} and \textit{Pedestrian}.
Following prior works~\cite{kim2021eagermot}, sequences \texttt{1, 6, 8, 10, 12, 13, 14, 15, 16, 18, 19} from the training set are designated as the validation set. 
Higher Order Tracking Accuracy (HOTA)~\cite{luiten2021hota} is adopted as the primary metric, providing a balanced evaluation of detection and association accuracy.
HOTA offers a unified evaluation of tracking quality by jointly penalizing localization errors, FP, FN, and IDS.
All evaluations are computed on the image plane.
We additionally evaluate Offline-Poly on the KITTI 3D benchmark with Scaled AMOTA (sAMOTA)~\cite{weng20203d}.

\subsection{Implementation Details}

\textbf{nuScenes.}
Our method is learning-free and implemented in Python using NumPy~\cite{van2011numpy}.
Leveraging the TBT framework, we conduct comparative evaluations across diverse trackers and detectors.
For ablation studies, we adopt the advanced online tracker Fast-Poly~\cite{li2024fast} with its default settings as the upstream tracker and CenterPoint~\cite{yin2021center} as the 3D detector.
Following prior works~\cite{ma2023detzero, CTRL, bitrack}, we extend the tracking outputs with forward-backward tracking, where the backward tracking reuses the forward configuration.
Across all experiments, we directly use the default configuration for open-source upstream trackers.
We then perform a linear search over the Offline-Poly parameters to maximize AMOTA on the validation set.
The resulting hyperparameters are subsequently applied to the test set.
Offline-Poly further incorporates category-specific techniques~\cite{li2023poly, li2024fast, wang2025mctrack}.
For the motion models used in the matching and refinement modules, we follow the configurations in~\cite{li2024fast}.
A comprehensive sensitivity analysis of the tuned hyperparameters is provided in Section~\ref{sec:sensitivity}.

\textbf{KITTI.}
We adopt the open-source trackers Fast-Poly~\cite{li2024fast} and MCTrack~\cite{wang2025mctrack} as upstream trackers, and VirConv~\cite{VirConv} as the 3D detector. 
MCTrack is used with its default configurations.
We adapt Fast-Poly to KITTI and perform a linear search to tune its parameters.
Bidirectional tracking is performed to expand the tracking results.
The hyperparameters of Offline-Poly are linearly searched on the validation set to optimize sAMOTA and then applied to the test set.
Following prior works~\cite{wang2025mctrack, bitrack}, evaluation is conducted on \textit{Car}.
The motion model used in matching and refinement is \textit{Constant Velocity}.

\begin{table*}[t!]
\centering
\caption{A comparison between Offline-Poly with other advanced methods on the nuScenes test set.
-F: Forward tracking. 
-B: Backward tracking.
Offline: Offline 3D MOT.
Structure: Low-level structure information (3D/2D bounding box~\cite{OTOP, kim2021eagermot, li2023poly, li2024fast, wang2025mctrack, li2023camo, lee2024dino}, map~\cite{OTOP}).
Semantic: High-context semantic information (feature map~\cite{wei2025bayesianmultiobjecttrackingneuralenhanced, li2023camo, lee2024dino}, object query~\cite{ding2024ada}).
}
\label{table:nu_test}
{\setlength{\tabcolsep}{1.0mm}{
\begin{tabular}{cccc|ccc|ccc}
\toprule
\multicolumn{1}{c}{\textbf{Method}} & \textbf{Detector(s)} & Input & \textbf{Offline} & \textbf{AMOTA}$\uparrow$ & \textbf{MOTA}$\uparrow$ & \textbf{AMOTP}$\downarrow$  & \textbf{IDS}$\downarrow$ & \textbf{FP}$\downarrow$ & \textbf{FN}$\downarrow$ \\ \midrule
OTOP~\cite{OTOP}  &  CenterPoint\cite{yin2021center}  & Structure   & \checkmark & 67.1 & 55.3 & 0.522    & 570    & 16778 & 22378  \\
EagerMOT~\cite{kim2021eagermot} & CenterPoint~\cite{yin2021center}\&Cascade R-CNN~\cite{cai2018cascade} & Structure & $\times$  & 67.7 & 56.8 &  0.550       & 1156   & 17705 & 24925  \\
CAMO-MOT~\cite{li2023camo} & BEVFusion~\cite{liu2023bevfusion}\&FocalsConv~\cite{chen2022focal} & Structure\&Semantic & $\times$  & 75.3  & 63.5 &  0.472   & 324   & 17269 & 18192  \\
Poly-MOT~\cite{li2023poly}   & LargeKernel-F\cite{chen2023largekernel3d}  & Structure   & $\times$ & 75.4 & 62.1 &\textbf{0.422}     & 292    & 17956 & 19673  \\
Fast-Poly~\cite{li2024fast}   & LargeKernel-F\cite{chen2023largekernel3d}  & Structure   & $\times$ & 75.8 & 62.8 &  0.479    & 326    & 17098 & 18415  \\ 
MCTrack~\cite{wang2025mctrack} & LargeKernel-F\cite{chen2023largekernel3d} & Structure & $\times$ & 76.3 & 63.4 & 0.445     & 242    &  19643  & 15996  \\	
DINO-MOT~\cite{lee2024dino} & Is-Fusion\cite{yin2024fusion} & Structure\&Semantic & $\times$ & 76.3 & 65.0 & 0.520    & 387    &  \textbf{15956}  & 17070  \\	
\midrule
Fast-Poly-F~\cite{li2024fast}  &  MV2DFusion\cite{mv2dfusion}  & Structure   & $\times$ & 76.3 & 64.5 & 0.505    & 266    & 20083 & \textbf{15812}  \\
Fast-Poly-B~\cite{li2024fast}  &  MV2DFusion\cite{mv2dfusion}  & Structure   & \checkmark & 76.7  & 64.9  & 0.489   & \textbf{203}    & 19342 & 15972  \\

\rowcolor{gray!25} \textbf{Offline-Poly (Ours)}   & MV-FP-F \& MV-FP-B   & Structure   & \checkmark & \textbf{77.6}  & \textbf{65.5} & 0.486     & 208    & 19098   & 15929   \\
\bottomrule
\end{tabular}}}
\vspace{-1.5em}
\end{table*}

\begin{table}[t]
\centering
\caption{A comparison of Offline-Poly with distinct detectors and trackers on the KITTI val set. 
$\dagger$: We evaluate methods with sAMOTA metric.
PG: PointGNN~\cite{shi2020point}.
VC: VirConv~\cite{VirConv}.
Bi: BiTrack~\cite{bitrack}.
MC: MCTrack~\cite{wang2025mctrack}.
FP: Fast-Poly~\cite{li2024fast}.
CAMO: CAMO-MOT~\cite{li2023camo}.
OP: Offline-Poly.
-F: Forward tracking. 
-B: Backward tracking.
Offline: Offline 3D MOT.
}
\label{table:scale_kitti}
\renewcommand{\arraystretch}{1}
\setlength{\tabcolsep}{1.0mm}{
\begin{tabular}{cccccc}
\toprule
\textbf{Index} & \textbf{Detector(s)} & \textbf{Tracker(s)} & \textbf{Offline} & \textbf{OP} & \textbf{sAMOTA}$\uparrow$ \\ \midrule
--         & VC        & Bi$\dagger$  & $\times$    & $\times$      & 91.44            \\
--         & VC        & Bi$\dagger$  & \checkmark     & $\times$      & 93.69            \\
--        & VC         & MC$\dagger$      & \checkmark & $\times$  & 94.20 \\
--         & PG        & CAMO  & $\times$     & $\times$      & 95.29           \\
\midrule
EXP1         & VC        & FP-F$\dagger$       & $\times$ & $\times$      & 95.51   \\ 
EXP2         & VC        & FP-B$\dagger$   & \checkmark    & $\times$      & 94.89            \\ 
\rowcolor{gray!25} EXP3         & VC        & FP-F\&FP-B     & \checkmark  & \checkmark       & \textbf{\textcolor{black}{97.69}} \textbf{{\scriptsize (+2.8)}} \\  
\midrule
EXP4         & VC         & MC-F$\dagger$       & $\times$   & $\times$   & 94.67 \\  
EXP5         & VC         & MC-B$\dagger$       & \checkmark  & $\times$    & 94.64  \\  
\rowcolor{gray!25} EXP6         & VC        & MC-F\&MC-B      & \checkmark   & \checkmark    & \textbf{\textcolor{black}{97.51}} \textbf{{\scriptsize (+2.9)}} \\  
\bottomrule
\end{tabular}}
\vspace{-1em}
\end{table}

\begin{table}[t]
        \centering
        \caption{A comparison of distinct local refinement implementations on the nuScenes val set.
        \textbf{KF}: Kalman Filter~\cite{kalman1960new}.
        \textbf{RTS}: Rauch-Tung-Striebel Smoother~\cite{rts}.
        \textbf{GPR}: Gaussian Process Regressor~\cite{bitrack}.
        \textbf{SWO}: Sliding-Window Optimization.
        }
        \label{table:local_refine}
        \renewcommand{\arraystretch}{1}
        \begin{tabular}{cccc}
        \toprule
        \multicolumn{1}{c}{\textbf{Refiner}} & \textbf{AMOTA}$\uparrow$ & \textbf{MOTA}$\uparrow$ & \textbf{AMOTP}$\downarrow$ \\ \midrule
         $\times$                       & 75.3      & 64.0     & 0.515  \\
         KF                       & 75.6      & 64.4     & 0.511  \\
         RTS       & 76.0      & 64.5     & 0.501  \\
         GPR     & 75.9      & 64.6     & \textbf{0.500}    \\
         \textbf{SWO (Ours)}      & \textbf{76.2}   & \textbf{64.8}   & 0.504  \\
        \bottomrule
        \end{tabular}
        \vspace{-2em}
\end{table}

\vspace{-1em}
\subsection{Comparative Evaluations}
\textbf{nuScenes.}
Table~\ref{table:scale_nusc} summarizes the scalability and generalization capability of Offline-Poly across heterogeneous detector-tracker combinations on the nuScenes validation set.
With the baseline detector CenterPoint and the advanced online tracker Fast-Poly, incorporating Offline-Poly consistently improves tracking accuracy, yielding AMOTA gains of +1.1\%, +1.2\%, and +2.5\% under forward, backward, and bidirectional settings, respectively. 
The bidirectional configuration achieves the highest overall performance (76.2\% AMOTA and 64.8\% MOTA), surpassing all existing online and offline methods by a clear margin.
When integrated with the pseudo-ground-truth multi-modal detector MV2DFusion, Offline-Poly further improves AMOTA by 1.5\% and MOTA by 1.3\%, demonstrating strong adaptability to detectors of varying modalities and quality.
Additional evaluations with the learning-based, multi-camera tracker ADA-Track exhibit even larger improvements (+2.7\% AMOTA and +2.9\% MOTA), highlighting the robustness and tracker-agnostic nature of Offline-Poly across data-driven trackers.
The results of EXP13 further demonstrate the advantages of multi-tracker integration. By combining appearance-based and geometry-based trackers, complementary spatio-temporal cues are leveraged, resulting in improved trajectory completeness (81.4\% AMOTA).
Consistent improvements across detectors with different modalities (LiDAR-only, Camera-only, LiDAR-Camera fusion) and tracking paradigms (TBD, TBA) further demonstrate the strong generalizability of Offline-Poly.
Collectively, these results substantiate the robustness and plug-and-play capability of Offline-Poly, establishing it as a unified and efficient component for both single- and multi-tracker 4DAL pipelines.


On the test set, we incorporate two upstream trackers: the bidirectional Fast-Poly tracking results combined with the MV2DFusion detector.
As shown in Table~\ref{table:nu_test}, by fusing multiple input, Offline-Poly delivers substantial improvements over each tracker. 
\textbf{We achieves state-of-the-art performance on the benchmark, attaining 77.6\% AMOTA among methods that rely exclusively on low-level structural information.}

\textbf{KITTI.}
As shown in Table~\ref{table:scale_kitti}, Offline-Poly demonstrates strong cross-dataset generalization and delivers superior tracking performance on longer, higher-frequency sequences.
On the validation set, Offline-Poly further exhibits robust transferability to diverse upstream trackers, yielding consistent performance gains (+2.8\% sAMOTA for Fast-Poly and +2.9\% sAMOTA for MCTrack).

On the KITTI test set, we employ a bidirectional Fast-Poly tracker with the VirConv detector as the upstream component.
\textbf{Offline-Poly establishes a new state-of-the-art on the KITTI tracking benchmark, achieving an HOTA of 83.00\% and outperforming all existing 3D MOT methods.}
Among offline trackers, Offline-Poly surpasses prior leading methods by a considerable margin (+2.6\% HOTA over RethinkMOT and +0.3\% HOTA over BiTrack).
Furthermore, Offline-Poly is learning-free and has minimal dependencies, benefiting from the TBT framework to perform offline tracking solely from existing tracking results.
In contrast, RethinkMOT requires additional training, whereas BiTrack depends on upstream tracking information and 2D segmentation.

\begin{table}[t]
        \centering
        \caption{A comparison of different global refinement implementations for \textit{Bus} on the nuScenes val set.
        \textbf{PC}: Position Correction.
        \textbf{SC}: Size Correction.
        }
        \label{table:global_refine}
        \renewcommand{\arraystretch}{1}
        \begin{tabular}{ccccc}
        \toprule
        \textbf{SC} & \textbf{PC} & \textbf{AMOTA}$\uparrow$ & \textbf{MOTA}$\uparrow$ & \textbf{ASE}$\downarrow$ \\ \midrule
         
         $\times$      & $\times$   & 88.0      & 78.7     & 0.171  \\
         Select All      & Center-Aligned   & 88.0    & 78.7     & 0.167  \\
         Select All      & Corner-Aligned   & 90.9     & 82.1     & 0.167  \\
    
         TopK    & Corner-Aligned     & \textbf{91.0}      & \textbf{82.3}      & \textbf{0.165}   \\

        \bottomrule
        \end{tabular}
        \vspace{-1em}
\end{table}

\begin{table}[t]
        \centering
        \caption{A comparison of different tracklet re-identification implementations (STWO in \cref{single_tracker}) on the nuScenes val set.
        \textbf{Motion}: Our proposed motion-based matching.
        \textbf{OTOP}: Re-ID module proposed in~\cite{OTOP}.
        Forward Fast-Poly using CenterPoint serves as the upstream tracker.
        }
        \label{table:single_stwo}
        \renewcommand{\arraystretch}{1}
        \begin{tabular}{cccc}
        \toprule
        \textbf{Matcher} & \textbf{Metric} & \textbf{MOTA}$\uparrow$ & \textbf{IDS}$\downarrow$ \\ \midrule
         
        OTOP      & \textit{NN}   & 64.1      & 367     \\
         
         Motion    & $\textit{gIoU}_{3d}$   & 64.7     & 380\\
         Motion    & $\textit{A-gIoU}_{3d}$~\cite{li2024fast}   & 64.8     & 380  \\
         Motion    & \textit{IoU}     & \textbf{65.0}      & \textbf{251}    \\
         
        \bottomrule
        \end{tabular}
        \vspace{-2em}
\end{table}

\subsection{Ablation Studies}
Ablation studies are conducted on the nuScenes validation set to quantitatively assess the contribution of each component within Offline-Poly. 
Unless otherwise specified, CenterPoint~\cite{yin2021center} is adopted as the baseline detector, and bidirectional Fast-Poly~\cite{li2024fast} serves as upstream trackers.

\textbf{The Effect of Tracklet-Level Pre-Processing.}
As shown in Table~\ref{table:nu_ablation}, applying a single filtering criterion (EXP1, EXP2) proves overly restrictive, resulting in lower recall and degraded overall performance. 
Incorporating both lifecycle and confidence constraints (EXP3) achieves a +0.3\% AMOTA improvement over the baseline by retaining reliable short tracklets and removing ghost ones.
This demonstrates that dual-criterion filtering better balances trajectory completeness and precision, providing cleaner inputs for downstream matching.

\textbf{The Effect of Single Tracker Matching and Fusion.}
As presented in Table~\ref{table:nu_ablation}, reorganizing the individual online tracking results with STWO (EXP8) and STW (EXP6) consistently improves performance.
This observation suggests that online trackers frequently suffer from fragmented or entangled trajectories, which can be effectively mitigated through our proposed offline matching strategy. 
We further find that the execution order of STWO and STW (EXP7) has negligible influence on the results, highlighting the decoupled nature of FN and FP matching and supporting the design of our unified tracker matching module.
Table~\ref{table:single_stwo} demonstrates that the tracklet re-identification (STWO) module is both modular and flexible within the TBT framework. 
The framework allows seamless substitution with a learning-based alternative (OTOP). 
Moreover, the proposed motion-based matching framework surpasses OTOP in performance, requires no training, and exhibits stronger generalization capability.

\begin{table}[t]
        \centering
        \caption{A comparison of distinct similarity metrics in multi-tracker matching on nuScenes val set.
        Eucl: Euclidean.
        }
        \label{table:multi_trackers}
        \renewcommand{\arraystretch}{1}
        \begin{tabular}{cccc}
        \toprule
        \textbf{Metric} & \textbf{AMOTA}$\uparrow$ & \textbf{MOTA}$\uparrow$ & \textbf{AMOTP}$\downarrow$ \\ \midrule
        \textit{Eucl}    & 73.3      & 63.3     & 0.549  \\
        $\textit{A-gIoU}_{3d}$~\cite{li2024fast}   & 75.6      & 64.7     & 0.513  \\
         $\textit{gIoU}_{3d}$   & 76.0      & 64.7     & 0.510  \\
         $\textit{IoU}_{3d}$     & \textbf{76.2}      & \textbf{64.8}     & \textbf{0.504}  \\
        \bottomrule
        \end{tabular}
        \vspace{-2em}
\end{table}

\begin{figure*}[t]
  \centering
   \includegraphics[width=1\linewidth]{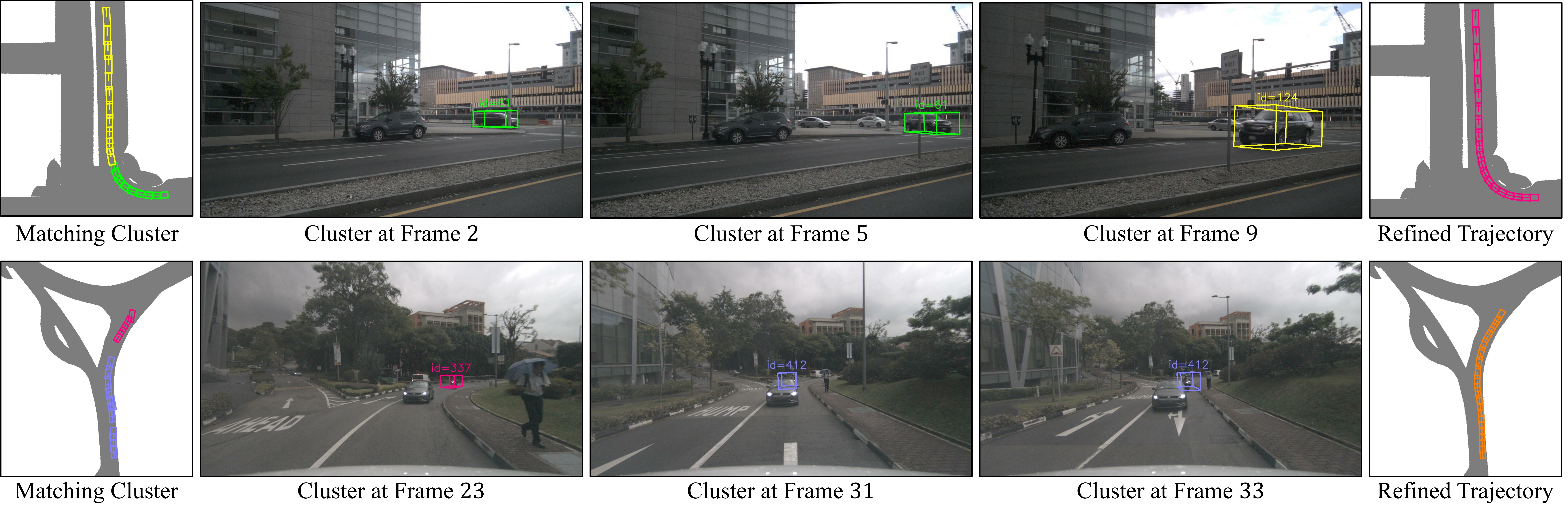}
   \caption{Visualization of the Single Trackers Matching (STM) module.
   The STM module re-identifies tracklet fragments caused by motion jitter (top) and occlusion (bottom) through the proposed motion-based matching framework, reducing IDS and FN.
}
   \label{viz:stm}
   \vspace{-1em}
\end{figure*}

\begin{figure*}[t]
  \centering
   \includegraphics[width=1\linewidth]{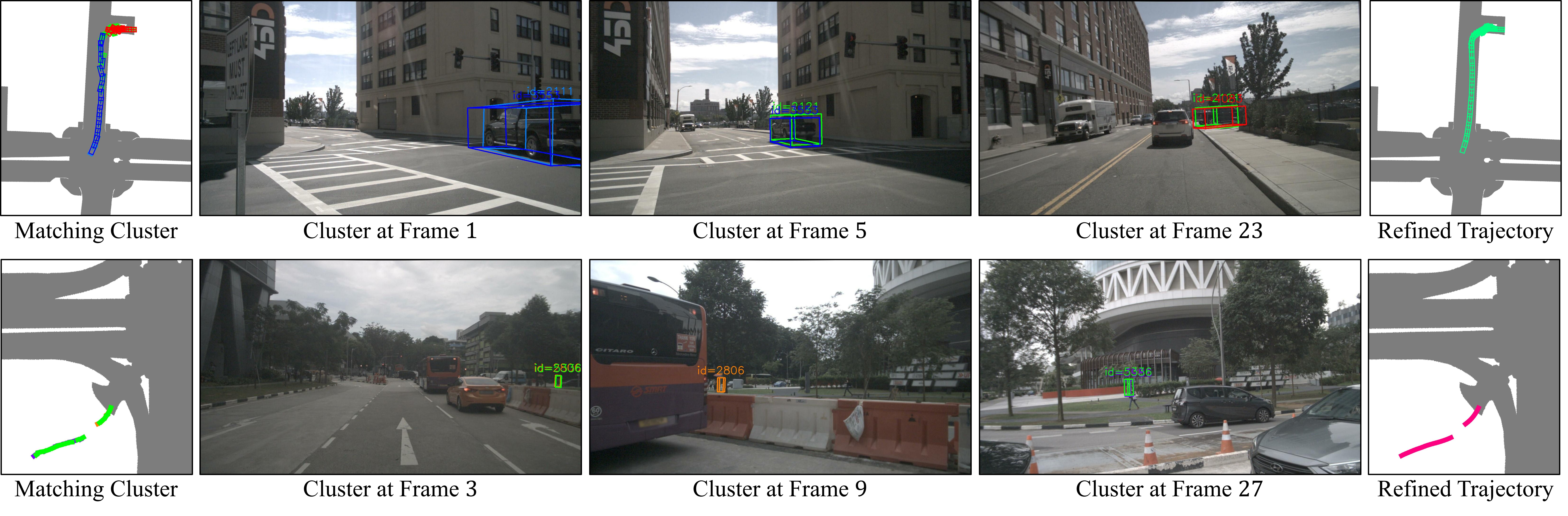}
   \caption{Visualization of the Multiple Trackers Matching (MTM) module. 
   The MTM module reconstructs more complete trajectories by fusing results from various upstream trackers.}
   \label{viz:mtm}
   \vspace{-1.5em}
\end{figure*}

\textbf{The Effect of Multiple Trackers Matching and Fusion.}
As illustrated in Table~\ref{table:scale_nusc}, integrating multiple upstream trackers (EXP5) yields a substantial performance improvement (up to 1.4\% AMOTA) over single-tracker optimization (EXP2).
The reduced AMOTP further indicates more accurate motion estimation.
Further increasing the number of integrated trackers (EXP13) continues to enhance accuracy (81.4\% AMOTA), demonstrating the scalability and complementary nature of multi-source fusion.
Table~\ref{table:multi_trackers} further confirms the generalization capability of this module, which performs robustly across different similarity metrics.
Among these, $\textit{IoU}_{3d}$ yields the best overall results, suggesting that stricter spatial affinity contributes to improved cross-tracker consistency.
Notably, as illustrated in Table~\ref{table:scale_nusc}, the enhanced observability provided by multi-tracker fusion also strengthens subsequent refinement, resulting in a +1.4\% AMOTA gain with multi-tracker input compared to +0.6\% with single-tracker input.

\textbf{The Effect of Multi-Perspective Trajectory Refinement.}
As shown in Table~\ref{table:nu_ablation}, the refinement module consistently enhances performance, yielding a +1.4\% AMOTA improvement with multi-tracker input and +0.6\% with single-tracker input. 
The comparison between EXP11 and EXP12 further indicates that global refinement provides a stronger initialization for subsequent local optimization.
As presented in Table~\ref{table:local_refine}, our motion-based refiner achieves higher accuracy and stronger physical interpretability than GPR, Kalman Filter (KF), and Rauch-Tung-Striebel (RTS) smoother, while requiring minimal parameter tuning.
As illustrated in Table~\ref{table:global_refine}, the corner-align strategy corrects positions via geometric inference, achieving a 3.4\% MOTA improvement for rigid objects. 
Global refinement based on high-confidence observations further enhances state estimation by effectively suppressing outliers.

\vspace{-0.5em}
\subsection{Hyperparameter Sensitivity Analysis}
\label{sec:sensitivity}
\begin{figure}
    \centering
    \includegraphics[width=1\linewidth]{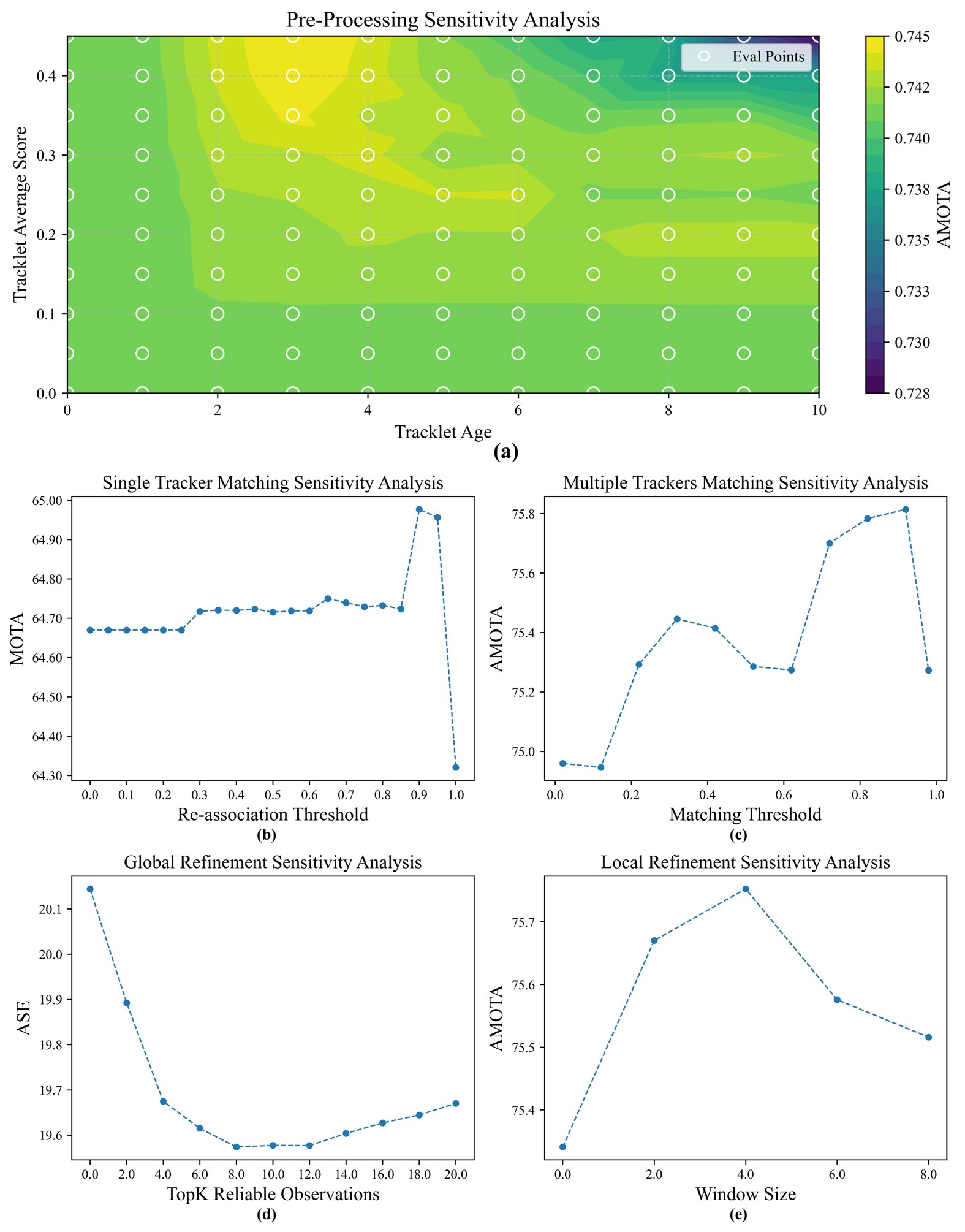}
    \caption{Performance comparison with identical key parameters across all categories.
    (a-b) employ CenterPoint with forward Fast-Poly.
    (c-e) use CenterPoint with bidirectional Fast-Poly.}
    \label{fig:hyper}
    \vspace{-1.5em}
\end{figure}

As shown in Fig.~\ref{fig:hyper}, we evaluate the robustness of Offline-Poly to its key hyperparameters.
With other parameters fixed, we respectively perform linear searches over tracklet age threshold $\theta_{a}$ and score threshold $\theta_{s}$ in pre-processing, tracklet re-identification threshold $\theta_{\text{blo}}$, the multi-tracker matching threshold $\theta_{\text{multi}}$, and the number of credible observations $K$ together with the sliding window size $M$ in refinement.

The lower-left region in Fig.~\ref{fig:hyper}(a) consistently outperforms the upper-right region, indicating that jointly removing short-lived and low-confidence trajectories substantially improves tracking performance.
As presented in Fig.~\ref{fig:hyper}(b), our proposed trajectory re-identification module (STWO) is generally insensitive to threshold variations and achieves stable gains across a wide range.
Optimal performance occurs under a moderately relaxed threshold (around 0.9), further verifying the robustness of explicit motion-model prediction over short-term horizons.
An excessively loose threshold ($=$1) leads to FP matching, degrading the system performance.
As illustrated in Fig.~\ref{fig:hyper}(c), the multi-tracker matching module favors a relatively relaxed threshold ($>$0.7), maximizing recall of geometrically consistent trajectories and reflecting strong complementarity among heterogeneous inputs.
For global refinement (Fig.~\ref{fig:hyper}(d)), reliable object size estimation requires a sufficient number of credible observations: too few ($<$4) lead to overfitting, whereas too many ($>$14) introduce outliers.
As displayed in Fig.~\ref{fig:hyper}(e), incorporating an appropriate near-future horizon improves sliding-window optimization.
Extending the prediction range beyond 4 seconds reduces accuracy due to motion-model drift.

\subsection{Visualizations}
In this section, we qualitatively demonstrate the effectiveness of the hierarchical matching and fusion mechanism.

\textbf{Single Tracking Matching (STM).}
As shown in Fig.~\ref{viz:stm}, the STM module leverages motion-based association to estimate potential continuity between tracklets, effectively matching and completing fragmented trajectories of the same object.
The top example illustrates a successful \textit{Car} tracklet association under high-speed cornering and motion jitter, while the bottom example demonstrates effective completion of a \textit{Moto} tracklet after long-term occlusion.

\textbf{Multiple Tracking Matching (MTM).} 
As shown in Fig.~\ref{viz:mtm}, we present two representative examples to illustrate the effectiveness of the MTM module. 
In the top example, irregular motion and occlusion in a turning scenario lead to frequent IDS. 
However, different trackers capture the same vehicle at different time intervals. 
By integrating their outputs, the MTM module effectively reduces these IDS occurrences and produces more complete trajectory observations.
The bottom example shows that for a \textit{Pedestrian} subject to long-term occlusions, the MTM module successfully fuses fragmented trajectory segments from multiple trackers, reconstructing more complete trajectories and improving recall.

\section{Conclusion}
In this paper, we present Offline-Poly, a polyhedral framework for offline 3D MOT built upon a novel TBT framework. 
We first introduce a track-centric offline tracking formulation that is fully decoupled from upstream detectors and trackers. 
Following the TBT framework, Offline-Poly constructs a four-stage pipeline that includes pre-processing, matching, fusion, and refinement to achieve robust tracking across multiple upstream trackers. 
Leveraging the ample computational resources and full-scene accessibility provided by offline settings, it filters ghost tracklets, re-identifies and reorganizes intra-tracker tracklets, re-associates cross-tracker observations of the same object, and refines the geometric and motion attributes of trajectories.
Extensive experiments on KITTI and nuScenes demonstrate that Offline-Poly is highly adaptable to a wide range of detectors and trackers and consistently delivers substantial performance gains. 
It achieves state-of-the-art results among both online and offline 3D MOT methods on these benchmarks. 
We release Offline-Poly as an open-source project to provide a strong baseline for future research.

\bibliographystyle{IEEEtran}
\bibliography{myref}

\newpage

 




\vfill

\end{document}